\documentclass{article}

\PassOptionsToPackage{numbers}{natbib}
\usepackage[main, final]{neurips_2026}

\usepackage[utf8]{inputenc} %
\usepackage[T1]{fontenc}    %
\usepackage{hyperref}       %
\usepackage{url}            %
\usepackage{booktabs}       %
\usepackage{amsfonts}       %
\usepackage{nicefrac}       %
\usepackage{microtype}      %
\usepackage{xcolor}         %
\usepackage{amsmath}
\usepackage{multirow}
\usepackage{graphicx}
\usepackage{colortbl}
\usepackage{tabularx}
\usepackage{bbm}
\definecolor{examgray}{gray}{0.92}
\usepackage{tcolorbox}
\usepackage{fontawesome5}
\newcommand{\hflogo}{\raisebox{-0.12em}{\includegraphics[height=0.95em]{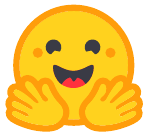}}}
\newcommand{\linkicon}[1]{\makebox[1.3em][l]{#1}}
\usepackage{listings}
\usepackage{enumitem}
\usepackage{subcaption}

\newtcolorbox{promptbox1}[1][]{
  colback=blue!3!white,
  colframe=blue!50!black!55!white,
  fonttitle=\bfseries\ttfamily\color{white},
  coltitle=white,
  title=#1,
  boxrule=0.5pt,
  arc=2pt,
  left=6pt, right=6pt, top=4pt, bottom=4pt,
}

\newtcolorbox{promptbox2}[1][]{
  colback=teal!3!white,
  colframe=teal!60!black!55!white,
  fonttitle=\bfseries\ttfamily\color{white},
  coltitle=white,
  title=#1,
  boxrule=0.5pt,
  arc=2pt,
  left=6pt, right=6pt, top=4pt, bottom=4pt,
}

\newcommand{\rev}[1]{{\color{blue}#1}}
\newcommand{\revstart}{\color{blue}}
\newcommand{\todo}[1]{{\color{red}\textbf{[TODO: #1]}}}
\renewcommand{\rev}[1]{#1}\renewcommand{\revstart}{}\renewcommand{\todo}[1]{}

\title{MedExAgent: Training LLM Agents to Ask, Examine, and Diagnose in Noisy Clinical Environments}

\author{%
  \rev{Yicheng Gao$^{1}$ \quad Xiaolin Zhou$^{2}$ \quad Yahan Li$^{1}$ \quad Yue Zhao$^{1}$ \quad Ruishan Liu$^{1}$} \\
  \rev{$^{1}$University of Southern California \quad $^{2}$Arizona State University} \\
  \rev{\texttt{\{gaoyiche,yahanli,yue.z,ruishanl\}@usc.edu}, \texttt{xzhou226@asu.edu}} \\
}

\begin{document}

\maketitle

\begin{abstract}
Real-world clinical diagnosis is a complex process in which the doctor is required to obtain information from both interaction with the patient and conducting medical exams. Additionally, the doctor needs to adapt to different patient personas, as well as noisy and incomplete information that can happen at any time during the process. However, existing benchmarks for medical LLMs and methods for automatic diagnosis largely simplify this process by reducing it to single-turn question answering, noise-free conversations, or sequential exam making, etc., ignoring the interactive and uncertain nature of clinical diagnosis. In this paper, we aim to address this gap by formalizing clinical diagnosis as a Partially Observable Markov Decision Process (POMDP) with three action types: questioning the patient, ordering medical exams as tool calls, and issuing a diagnosis. We also introduce a systematic noise model comprising seven patient noise types and three exam noise types. Using our proposed environment, we train an effective diagnosis agent, \textbf{MedExAgent}, through a two-stage pipeline that first performs supervised finetuning on synthetic conversations structured after the Calgary-Cambridge model for clinical interviews, and then applies DAPO to optimize a composite reward capturing diagnostic accuracy, tool call quality, and exam cost including financial cost and patient discomfort. Through extensive experiments and ablation studies, we demonstrate that MedExAgent achieves diagnostic performance comparable to larger models while maintaining cost-efficient examination strategies.

\vspace{3pt}
{\small
\linkicon{\faGlobe}\textbf{Project page:} \href{https://endlesscg.github.io/medexagent/}{\texttt{endlesscg.github.io/medexagent}}\\
\linkicon{\faGithub}\textbf{Code:} \href{https://github.com/EndlessCG/medexagent}{\texttt{github.com/EndlessCG/medexagent}}\\
\linkicon{\hflogo}\textbf{Model:} \href{https://huggingface.co/medagent/MedExAgent-8B}{\texttt{huggingface.co/medagent/MedExAgent-8B}}\\
\linkicon{\faDatabase}\textbf{Data:} \href{https://huggingface.co/datasets/medagent/MedExAgent-Data}{\texttt{huggingface.co/datasets/medagent/MedExAgent-Data}}\par}
\end{abstract}

\begin{figure}[tbh]
    \centering
    \resizebox{\linewidth}{!}{%
      \includegraphics[trim={0bp 340bp 510bp 0bp}, clip]{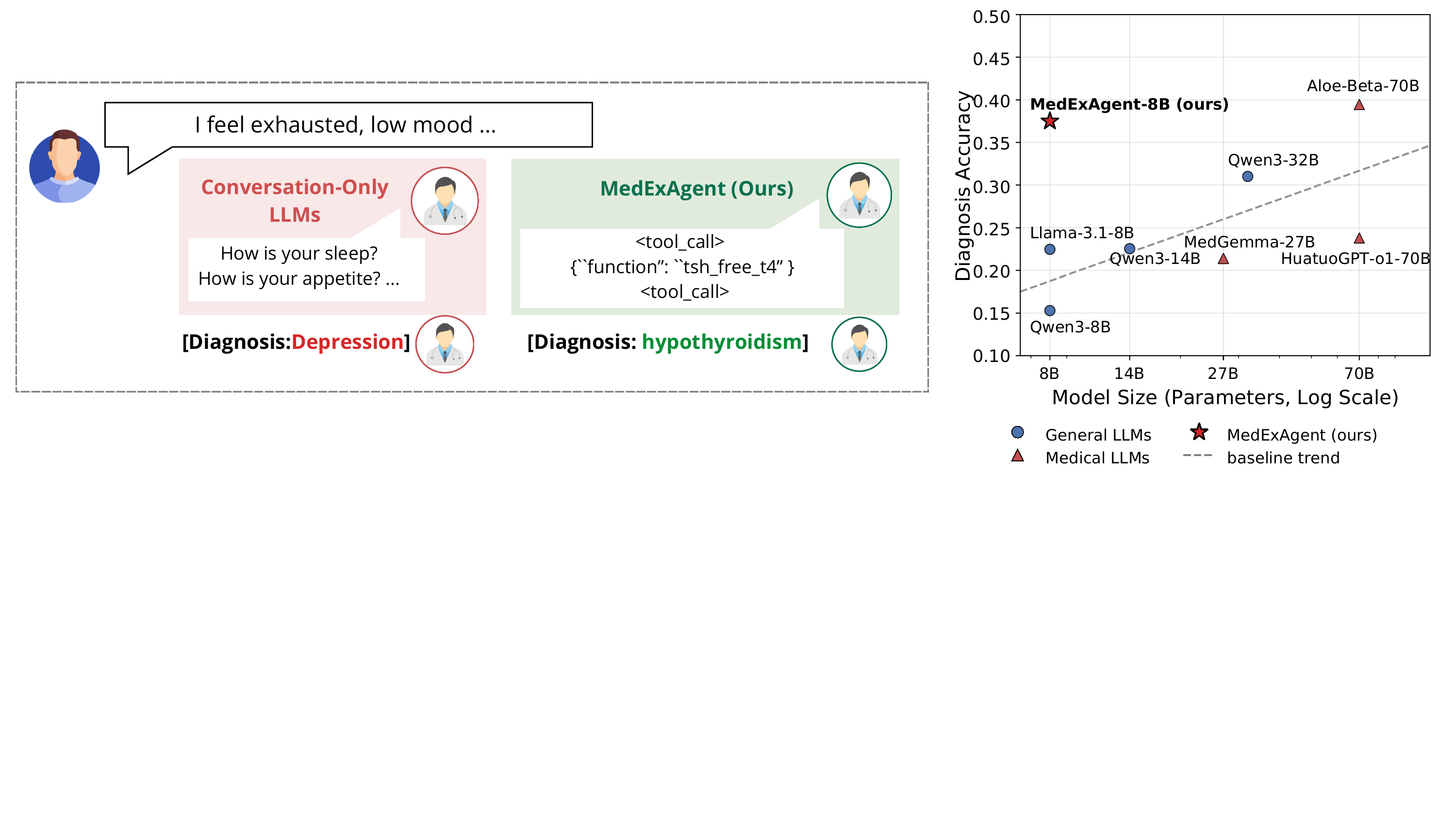}%
      \includegraphics[height=470bp]{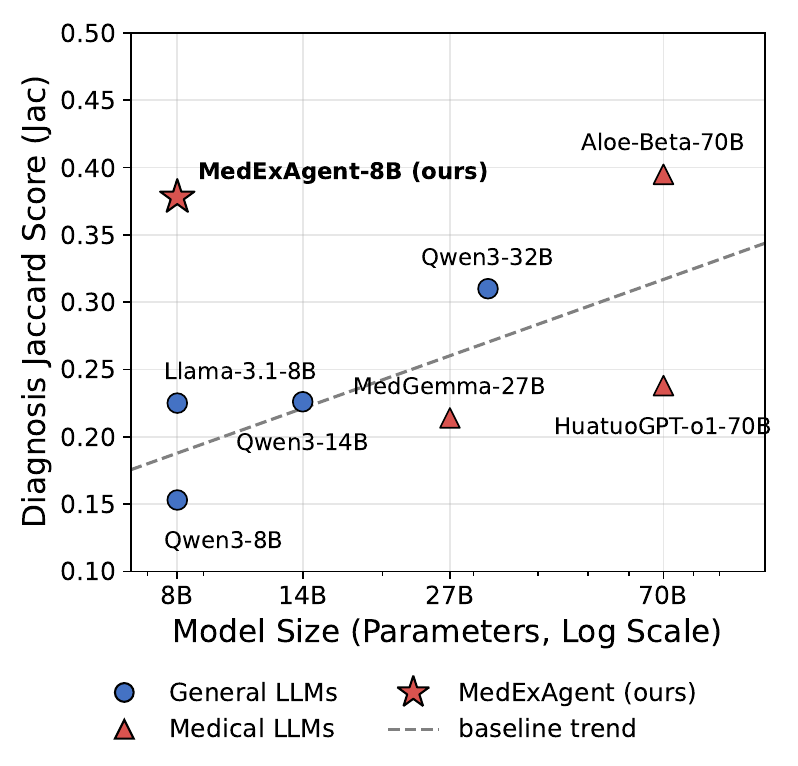}%
    }
    \caption{\textbf{MedExAgent overview.} (a) Tool-augmented diagnostic reasoning enables correct diagnoses where conversation-only baselines fail. (b) On the OOD test set AgentClinic-MedQA~\cite{schmidgall2024agentclinic}, Our 8B agent matches much larger baselines on \rev{Jac}.}
    \label{fig:placeholder}
\end{figure}

\begin{center}
\fbox{\parbox{0.95\linewidth}{\small\textbf{Disclaimer:} MedExAgent is a research prototype. It must not be used for medical advice or patient care. 
}}
\end{center}

\section{Introduction}

Diagnostic errors contribute to substantial patient harm each year, and access to timely, high-quality diagnostic care remains uneven across populations and geographies \cite{burden_of_serious_harms_from_diagnostic_error_in_usa, AAMC_physician_workforce_projections, bridging_the_health_care_gap_in_rural_populations}. These pressures have motivated longstanding interest in computational support for clinical reasoning \cite{an_oerview_of_clinical_decision_support_systems}, and recent advances in large language models (LLMs) have brought this goal within closer reach. Medical LLMs now show strong performance on knowledge-intensive benchmarks such as MedQA \cite{medqa} and MedMCQA \cite{medmcqa}, and a growing line of medical specialists \cite{meditron, biomistral, sellergren2025medgemma, huatuogpto1} are trained for these tasks; more recent diagnostic evaluations move closer to clinical use by asking models to reason over complete clinical cases \cite{DiagnosisArena, use_of_gpt4_to_diagnose_complex_clinical_cases, comparative_analysis_of_multimodal_llm_performance_on_clinical_vignette_questions}. 
Across this landscape, the model is handed a complete case description and asked to reason over it; however, in clinical practice, the case must be constructed through interaction with the patient. Building a deployable \emph{diagnostic agent}, one that constructs the case itself, in real clinical conditions, is the subject of this work.

Clinical diagnosis as practiced differs from these benchmarks in three respects. It is \emph{sequential}: a clinician begins with a chief complaint, asks questions to refine a working differential, orders exams as new evidence is needed, and commits to a diagnosis \cite{balogh2015diagnostic}. It is \emph{noisy}: patients misreport symptom location or severity \cite{Redelmeier_Tu_Schull_Ferris_Hux_2001_problem_of_clinical_judgement_obtaining_a_reliable_past_medical_history, Lackner_Jaccard_Keefer_Firth_Carosella_Sitrin_Brenner_2014_the_accuracy_of_patient_reported_measures_for_gi_symptoms}, and laboratory results can be contaminated, partially missing, or ambiguous \cite{Shreffler_2023_diagnostic_testing_accuracy}. And it is \emph{costly}: Exams range from a routine blood draw to an invasive biopsy, with corresponding differences in financial cost \cite{overtreat_in_the_us, Ji_Carin_2007_cost_sensitive_feature_acquisition} and patient discomfort that clinicians weigh against expected diagnostic yield \cite{lu2024benefits}. Static benchmarks capture none of these properties.

Recent work has begun to move LLMs into interactive settings, but each effort handles only part of the picture. Benchmarks evaluate patient questioning \cite{li2024mediq, schmidgall2024agentclinic} or EHR tool-calling \cite{jiang2025medagentbenchrealisticvirtualehr}; specialist agents are trained for conversational history-taking \cite{amie, amie2} or for exam ordering \cite{diagagent}. Three properties of clinical practice consequently remain absent from the training signal of any single agent. First, no existing system optimizes a single agent over the combined action space of questioning, exam ordering, and diagnosis within one episode. Second, patient and exam observations are treated as clean rather than noisy. Third, the cost of investigation is rarely part of the objective. Whether a single agent can be trained to handle sequential interaction, noise, and cost together, and what such training requires, remains open.

We address this question along two axes: an environment that exposes the agent to all three properties, and a training recipe that teaches it to act within them. We formalize interactive diagnosis under observational uncertainty as a Partially Observable Markov Decision Process (POMDP). The action space unifies free-form patient questioning, exam ordering as tool calls, and final diagnosis under a single agent, closing the interleaving gap. Observational noise is injected stochastically into the observation function via a systematic noise model (seven patient noise types, three exam noise types), with diverse patient personas drawn from PatientSim \cite{patientsim}. Per-exam cost is grounded in CMS fee schedules \cite{cms_pfs_rvu26a, cms_clfs_2025q4} for financial cost and in the access-and-invasiveness taxonomy of Whiteley et al.\ \cite{Whiteley2024AccessInvasiveness} for patient discomfort, so that an exam's cost reflects what it would impose on a real patient.

Within this environment we train \textbf{MedExAgent}, an 8B-parameter agent built on Meditron3-8B \cite{meditron}.  We draw on the Calgary--Cambridge model of medical interviewing \cite{kurtz1996calgary, calgary3}, a widely adopted framework in medical education that decomposes a clinical encounter into five stages: (1) initiating the session, (2) gathering information, (3) physical examination, (4) explanation and planning, and (5) closing. Doctor--patient conversations are synthesized following these stages and used as supervision. Training proceeds in two stages: supervised fine-tuning on these conversations, followed by DAPO \cite{yu2025dapoopensourcellmreinforcement} reinforcement learning against a composite reward over diagnostic accuracy, tool-call quality, and per-exam cost. Across in-distribution and out-of-distribution benchmarks, MedExAgent-8B matches or exceeds far larger baselines on diagnostic accuracy while maintaining cost efficiency.

Concretely, we make three contributions:

\begin{enumerate}
\item A POMDP environment for interactive clinical diagnosis that unifies patient questioning, exam ordering, and diagnosis, with patient and exam noise injected into the observation function and per-exam cost reflecting real-world financial burden and patient discomfort.
\item A training recipe that combines supervised fine-tuning on Calgary--Cambridge-structured doctor--patient conversations with reinforcement learning against a composite reward over diagnostic accuracy, tool-call quality, and per-exam cost. 
\item An open 8B model showing that a small specialist trained in this environment matches or exceeds much larger baselines on diagnostic accuracy while maintaining cost-efficient examination strategies.
\end{enumerate}

\section{Related Works}

\subsection{Benchmarks and Environments for Interactive Diagnosis} Following the recent upsurge of LLM-based agents, a line of work aims to bridge the gap between text-based LLMs and interactive real-world clinical environments. MediQ~\cite{li2024mediq} introduces a benchmark allowing the LLM to communicate with the patient. Results demonstrate that giving LLMs the flexibility to inquire the patient can boost performance significantly, highlighting the importance for agent-environment interaction. Similarly, AgentClinic~\cite{schmidgall2024agentclinic} proposes a multi-agent framework to assess LLM's diagnosis performance in an environment allowing multi-modal information gathering. MedAgentBench~\cite{jiang2025medagentbenchrealisticvirtualehr} further evaluates LLMs in EHR-based clinical tasks such as patient information retrieval, lab test ordering, etc. However, existing environments and benchmarks often assume that the response to the action of LLMs are accurate, which ignores the noises that are common in real world. In fact, a large-scale study of 5,900 patients found only 5.5\% complete agreement between self-reported data and EHR records~\cite{pcori}. Our work addresses this gap by systematically modeling patient noise, exam noise, and  patient personalities in our environment.

\subsection{LLM-based Systems for Clinical Diagnosis} Previous works in applying LLMs to conduct clinical diagnosis can be broadly split into two categories: multi-agent systems and finetuned expert models. Methods involving multi-agent systems typically assign different roles to a set of LLMs and design a workflow to orchestrate the diagnosis process. MAI-DxO~\cite{mai_dxo_sdbench} uses five LLMs to each role-play a step in the workflow (e.g. planning, hypothesis generation, cost management). MedAgents~\cite{medagents} and MDAgents~\cite{MDAgents} assign each LLM to a medical field (e.g. radiologist, oncologist, surgeon). These methods require extensively calling APIs of closed-source models, raising concerns about patient privacy which is highly sensitive in the field of medicine. Another line of work aims to finetune expert models for interactive clinical diagnosis. AMIE~\cite{amie, amie2} and ClinDiag-GPT~\cite{Chen2026} both train closed-source expert models for conversation-based information acquisition. DiagAgent~\cite{diagagent} instead trains an agent to predict medical examinations for a given patient. Both directions only model part of the complete diagnosis procedure (either conversation or exam selection), which can restrict effectiveness and real-world authenticity.

\section{Problem Formulation}\label{sec:problem_formulation}

In this paper, we study the problem of automatic diagnosis with medical exams as tool calls. We formalize this problem as a POMDP $(\mathcal{S}, \mathcal{A}, \textit{T}, \Omega, \textit{O}, \textit{R})$. The state space $\mathcal{S}=\mathcal{D}$ is a set of all possible diseases, from which a random disease $d$ is sampled at the beginning of each episode as the ground truth disease the patient has. The action space $\mathcal{A}$ contains three types of actions the agent can take in each timestep $t$:  $\mathcal{A}=\mathcal{A}_{\text{ask}}\cup\mathcal{A}_{\text{exam}}\cup\{a_{\text{dx}}\}$. If $a_t\in\mathcal{A}_{\text{ask}}$, the agent asks the patient a question; if $a_t\in\mathcal{A}_{\text{exam}}$, the agent chooses one exam from a predefined available exam list; if $a_t=a_{\text{dx}}$, the agent gives a final diagnosis $\hat{d}\in\mathcal{D}$.  We assume that the patient's disease does not change or progress during the episode. This indicates that the transition function is deterministic: $T(s' \mid s, a) =\mathbbm{1}(s'=s)$, where $\mathbbm{1}$ is the indicator function. We detail other parts of the POMDP as follows.

\subsection{Noise Modeling}\label{sec:noise_modeling}
In our POMDP definition, the observation space $\Omega$ contains two types of observations: $\Omega=\Omega_{\text{conv}}\cup\Omega_{\text{exam}}$, where $\Omega_{\text{conv}}$ represents all possible responses of the patient and $\Omega_{\text{exam}}$ represents all possible exam results. At timestep $t$, the observation $\omega_t\in\Omega_{\text{conv}}$ if and only if the action $a_t\in\mathcal{A}_{\text{ask}}$, and $\omega_t\in\Omega_{\text{exam}}$ if and only if the action $a_t\in\mathcal{A}_{\text{exam}}$. 

In real world, a medical diagnosis session often contains noises. For example, the patient can mistakenly report the region of a pain (e.g. stomach pain $\rightarrow$ liver pain), and a medical exam can misreport certain symptoms (e.g. contaminated samples result in inaccurate blood test results). In order to reflect this, we model such noises in the observation function $O(\omega \mid d, a)$. We consider seven types of patient noises and three types of exam noises detailed in Table~\ref{tab:noise_types}. Each type of noise is applied independently with a fixed probability $p_{\text{conv}}$ or $p_{\text{exam}}$ on patient responses or exam results. \rev{Our noise taxonomy is derived from common error classifications in medical research, which repeatedly distinguish missing, incorrect or misinterpreted, and uncertain information. For example, Schiff et al.~\cite{schiff2009diagnostic} separate failure to elicit information from inaccurate or misinterpreted information, Kim and Mansfield~\cite{kim2014fool} distinguish under-reading and faulty reasoning in radiology, and Callen et al.~\cite{callen2020between} study uncertain findings in radiology reports. These correspond to the omission, substitution, and ambiguity noises we use. Our patient noises also align with recent taxonomies of difficult patient behaviors~\cite{li2026beyond, luo2026meddialbench}.}

\subsection{Reward Design}\label{sec:reward}
Our reward function $R$ is the total episode reward containing three parts: diagnosis reward, tool call matching reward, and exam cost reward. The \textbf{diagnosis reward} $R_{\text{dx}}$ measures the agreement between the predicted diagnosis and the ground truth diagnosis. Since a single case may involve multiple conditions (e.g., ``arm fracture with mild infection'') and equivalent conditions can be expressed differently (e.g., ``influenza'' and ``flu''), we use an LLM judge (\texttt{gpt-4.1-mini}) to extract the set of conditions from both the prediction and the ground truth, match semantically equivalent conditions, and compute a Jaccard score:
\begin{equation}\label{eq:r_dx}
R_{\text{dx}} = \frac{|\mathcal{G} \cap \mathcal{P}|}{|\mathcal{G} \cup \mathcal{P}|}
\end{equation}
where $|\mathcal{G}|$ and $|\mathcal{P}|$ denote the number of conditions identified by the judge in the ground truth and prediction, respectively, and $|\mathcal{G} \cap \mathcal{P}|$ is the number of matched conditions. The prompt templates are shown in Appendix~\ref{sec:prompt-templates}. We verify the accuracy of the LLM judge in Appendix~\ref{sec:llm_judge_verify}.

The \textbf{tool call reward} $R_{\text{tool}}$ considers the overlap between the tool calls of the predicted trajectory and the ground truth trajectory. We follow ToolRL\cite{qian2025toolrlrewardtoollearning} and use a three-level matching reward:
\begin{equation}
 R_{\text{tool}} =
  \frac{J_{\text{name}} + \sum_{i=1}^{|\mathcal{T}_{\mathcal{G}}|} \left( J_{\text{param}}^{(i)} + V^{(i)} \right)}
       {1 + \sum_{i=1}^{|\mathcal{T}_{\mathcal{G}}|} \left( 1 + |\theta_i^{\mathcal{T}_{\mathcal{G}}}| \right)}.
\end{equation}
where $J_{\text{name}} = \frac{\sum_{x} \min(c_{\mathcal{T}_{\mathcal{P}}}(x), c_{\mathcal{T}_{\mathcal{G}}}(x))}{\sum_{x}\max(c_{\mathcal{T}_{\mathcal{P}}}(x), c_{\mathcal{T}_{\mathcal{G}}}(x))}$
is the multiset Jaccard over tool names ($c_{\mathcal{T}_{\mathcal{P}}}, c_{\mathcal{T}_{\mathcal{G}}}$ are name counts in predicted/ground-truth),
$J_{\text{param}}^{(i)}$ is the set Jaccard over parameter names for the $i$-th greedy-matched tool
pair,
$V^{(i)} = \sum_{k \in \theta_i^{\mathcal{T}_{\mathcal{G}}}} \mathbbm{1}[\theta_i^{\mathcal{T}_{\mathcal{G}}}(k) =
\theta_i^{\mathcal{T}_{\mathcal{P}}}(k)]$ is the number of exact parameter value matches,
and $|\theta_i^{\mathcal{T}_{\mathcal{G}}}|$ is the number of parameters in the $i$-th ground-truth tool. We modify
the ToolRL reward function by replacing the  set Jaccard
score in $J_{\text{name}}$ with a multiset (count-aware) Jaccard,  which penalizes the agent for calling the same tool multiple times unnecessarily.

The \textbf{exam cost reward} aims to prevent the agent from calling unnecessarily costly exams. We consider two aspects: financial cost of the exam, and patient discomfort induced by the exam. For each aspect, we classify the cost into three tiers on a three-point ordinal scale (Low=1, Medium=2, High=3). For financial cost, we follow CMS Physician Fee Schedule RVU values (2026)~\citep{cms_pfs_rvu26a} for procedures and imaging, and CMS Clinical Laboratory Fee Schedule rates (2025Q4)~\citep{cms_clfs_2025q4} for laboratory tests. 
For patient discomfort, we follow \cite{Whiteley2024AccessInvasiveness} and distinguish procedures by varying degrees of penetration. Specifically, non-invasive exams (e.g., external physical exam, X-ray) are tier 1, minimally invasive exams involving superficial penetration (e.g., venipuncture, swabs, standard endoscopy) are tier 2, and invasive procedures (e.g., biopsy, catheterization, surgical exploration) are tier 3. Lab tests inherit the discomfort tier of their collection method. We sum both types of costs for all unnecessary exams and divide by 6, which is the maximum possible cost value for a single exam.

Finally, we combine the three reward terms by calculating a weighted sum:

\begin{equation}\label{eq:reward-function}
    R(d, a_{1:T}) = R_{\text{dx}} + w_{\text{tool}}R_{\text{tool}} - w_{\text{cost}}R_{\text{cost}}.
\end{equation}

where $w_{\text{tool}}$ and $w_{\text{cost}}$ are hyperparameters.

\section{Dataset and Environment Construction}\label{sec:env_construction}

Training the agent against the POMDP defined in Section~\ref{sec:problem_formulation} requires two ingredients: a corpus of doctor--patient conversations for supervised fine-tuning, and an interactive environment for reinforcement learning. We describe both in this section.

\subsection{Data Sources}\label{sec:data_sources}

\paragraph{DDxPlus.}
DDxPlus~\cite{fansi2022ddxplus} is a large-scale synthetic differential diagnosis dataset of roughly 1.3M patients, covering 49 pathologies and 110 symptoms. DDxPlus is designed to reproduce authentic joint distributions over demographics, pathologies, and symptoms in the U.S. rather than a uniform or adversarial mix. This makes DDxPlus well-suited for optimizing an agent towards addressing ``typical'' patients, reflecting realistic base rates. However, its 49-pathology label space is narrow by design and under-represents rarer conditions.

\paragraph{PMC-Patients-v2.}
PMC-Patients-v2~\cite{zhao2023large} is derived from real clinical case reports published in PubMed Central, comprising approximately 250{,}294 patient records. Because the editorial threshold for publishing a case report favors unusual, diagnostically challenging, or instructive presentations, PMC-Patients-v2 is naturally enriched for rare diseases, atypical manifestations of common diseases, and multi-condition cases, complementing our dataset with long-tail cases. 

\paragraph{AgentClinic-MedQA.} 
AgentClinic~\cite{schmidgall2024agentclinic} is a benchmark containing OSCE-style structured patient profiles, including medical history, symptoms, and test results of the patients. We use the MedQA subset of the benchmark since other subsets contain multimodal entries. AgentClinic-MedQA contains 201 cases grounded in the USMLE questions from the MedQA dataset. These cases are used as our Out-Of-Distribution (OOD) test set and held out from the training set.

\subsection{SFT Conversation Generation}\label{sec:conversation_generation}

\begin{figure*}
    \centering
    \includegraphics[width=\linewidth, trim={0.5cm 2.5cm 0.3cm 0.5cm}, clip]{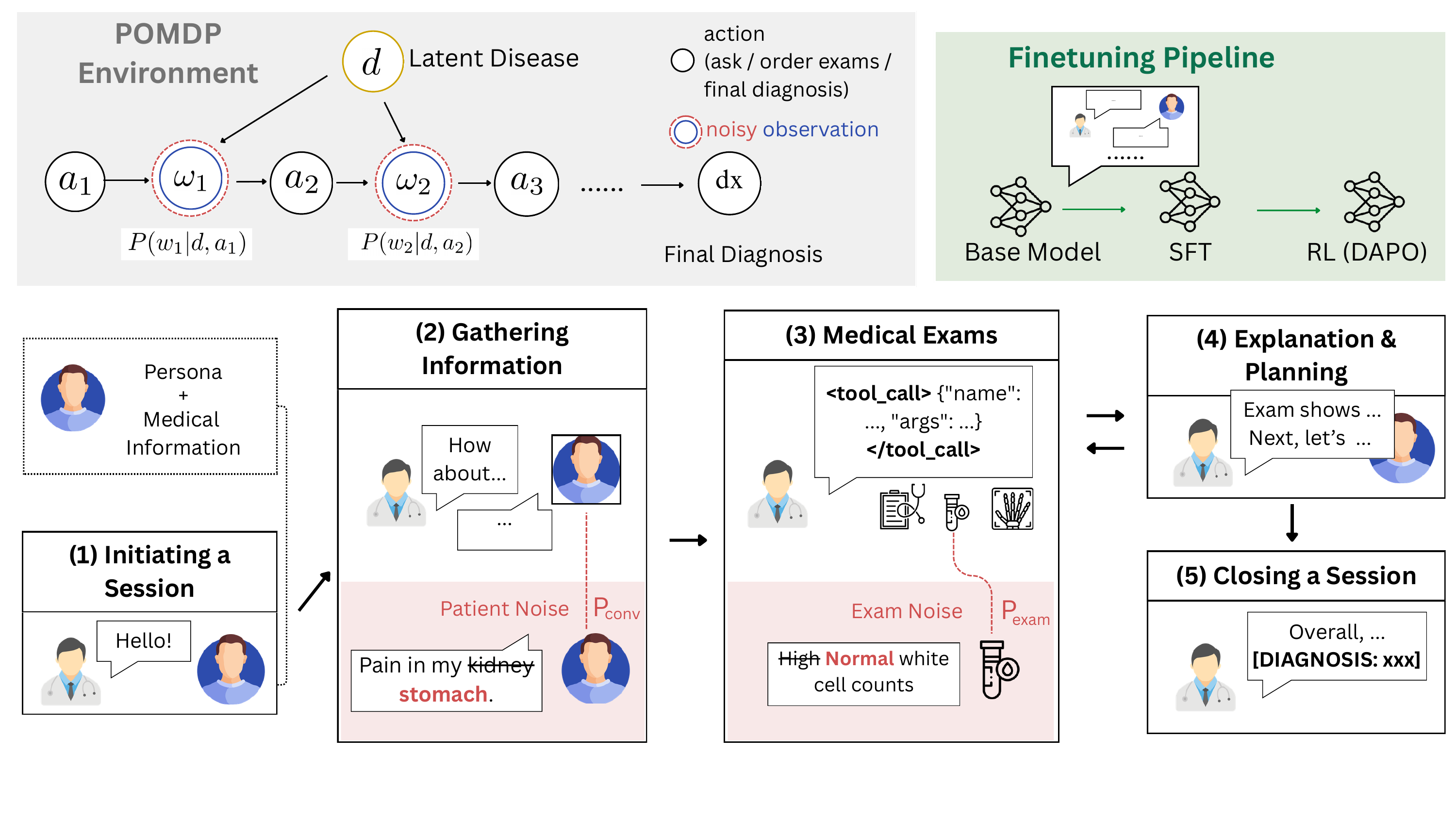}
    \caption{\textbf{Method overview.} Top left: The interactive diagnosis POMDP, with latent disease $d$ generating noisy observations $\omega_t$ from agent actions $a_t$. Top right: Two-stage finetuning (SFT then DAPO RL). Bottom: Doctor--patient conversations follow the five-stage Calgary--Cambridge model, with patient and exam noise injected at rates $p_{\text{conv}}$ and $p_{\text{exam}}$.}
    \label{fig:ccmodel}
\end{figure*}

We now describe our pipeline to generate conversations for the SFT stage of our training pipeline. To begin with, we subsample DDxPlus and PMC-Patients-v2 to form a training set. We sample a total of 5,000 samples for SFT, 20,000 for RL, and 2,000 for test set, each evenly distributed between DDxPlus and PMC-Patients-v2.  From each case, we construct a data point containing patient demographics, medical history, self-reported symptoms, exams conducted, exam results, and the final diagnosis.

We generate conversations following the five-stage Calgary--Cambridge model~\cite{kurtz1996calgary} for medical interviews, using two LLMs to simulate the doctor and patient. As shown in Figure~\ref{fig:ccmodel}, each conversation can be split into five stages: initiating a session, gathering information, medical exams, explanation and planning, and closing a session. The doctor agent greets the patient and gathers self-reported symptoms, then enters a loop of ordering exams via tool calls and explaining results until it has enough evidence to commit to a diagnosis, summarize the session, and close. A stage-specific prompt is provided to each LLM at every turn (Appendix~\ref{sec:prompt-templates}).

In order to better reflect patient behavior in the real world, we randomly sample one patient persona for each conversation. We sample from a total of 36 patient personas proposed in PatientSim~\cite{patientsim}, excluding the ``highly confused'' persona where the patient is unable to communicate. \rev{The realism of these personas has been verified through expert evaluation in PatientSim, and we directly apply their taxonomy in our environment. In Appendix~\ref{sec:persona_ablation}, we show that the personas produce the intended differences in patient communication but do not significantly affect diagnostic performance. Therefore, we use them to increase the diversity of our patient simulator.} Additionally, after generating the conversations, we apply the patient and exam noises described in Section~\ref{sec:noise_modeling}.

\subsection{RL Environment Construction}\label{sec:rl_environment}

We instantiate the POMDP defined in Section~\ref{sec:problem_formulation}. Each patient profile from Section~\ref{sec:data_sources} defines one episode instance with: (i) ground-truth disease $d$; (ii) demographics, medical history, and self-reported symptoms; (iii) an exam map $\mathcal{M}$ storing canonical findings for each recorded exam; and (iv) an available tool set $\mathcal{A}_{\text{exam}}$, consisting of ground-truth exams plus randomized distractors sampled from a predefined taxonomy.

The trained model is initialized with a system prompt specifying demographics, medical history, OpenAI-style schemas of $\mathcal{A}_{\text{exam}}$, and instructions for the conversation. Detailed prompt templates are shown in Appendix~\ref{sec:system_prompt_templates}. The trained model starts the conversation by greeting the patient as instructed in the system prompt. At each turn, the trained model emits one of three action types:
\begin{itemize}
    \item $a_{\text{ask}}$: a free-form question forwarded to a patient simulator (an LLM conditioned on the case profile and a sampled persona same as in Section~\ref{sec:conversation_generation}) that returns an in-character reply.
    \item $a_{\text{exam}}$: a \texttt{<tool\_call>} block parsed against $\mathcal{M}$. Exact matches return canonical findings; available but non-ground-truth tools return ``No significant findings''; tools outside $\mathcal{A}_{\text{exam}}$ return ``This exam is not available''. 
    \item $a_{\text{dx}}$: a free-form text ending in \texttt{[DIAGNOSIS: \dots]}, which terminates the episode and triggers reward computation (Equation~\ref{eq:reward-function}).
\end{itemize}

Additionally, for each episode, we randomly sample patient and exam noises with probability $p_{\text{conv}}$ and $p_{\text{exam}}$ as described in Section~\ref{sec:noise_modeling}. When a patient round is sampled for noise injection, we append a noise-specific prompt to the patient's system prompt. For exam noises, we use a deterministic rule to post-process exam findings. More details are described in Appendix~\ref{sec:noise_injection_details}.

\section{MedExAgent Training and Evaluation}\label{sec:training_evaluation}
\subsection{Finetuning Pipeline}

We use the synthesized dataset described in Section~\ref{sec:conversation_generation} to finetune an LLM-based agent. We use Meditron3-8B~\cite{meditron} as our base model since it's the state-of-the-art medical LLM with around 8B parameters, which allows for easier deployment in medical environments with lower compute resources. We use a two-stage pipeline with an SFT stage followed by a RL stage. In the SFT stage, we finetune the model with the generated conversations. In the RL stage, we apply DAPO~\cite{yu2025dapoopensourcellmreinforcement} to train the model using the extracted patient information described in Section~\ref{sec:conversation_generation}. For detailed hyperparameter settings, please refer to Appendix~\ref{sec:hyperparameters}.

\subsection{Evaluation Pipeline}\label{sec:evaluation_pipeline}

\paragraph{Test sets.}
We evaluate our trained model and baseline models on three benchmarks: the test splits of DDxPlus and PMC-Patients-v2, and AgentClinic-MedQA as an out-of-distribution (OOD) benchmark. For each evaluation set, we provide the model with the same system prompt used during SFT conversation generation, including instructions for the full diagnostic procedure and the list of available medical exams. We use 1000 samples each for DDxPlus and PMC-Patients-v2, and all (201) samples for AgentClinic-MedQA.

\paragraph{Baseline models.}
We compare MedExAgent against a total of seven baseline models. This include four general-purpose LLMs: Qwen3-8B, Qwen3-14B, Qwen3-32B~\cite{yang2025qwen3}, Llama-3-8B-Instruct~\cite{grattafiori2024llama}, and three medical LLMs: Aloe-Beta-70B~\cite{garcia2025aloe}, MedGemma-27B-text-it~\cite{sellergren2025medgemma} and HuatuoGPT-o1-70B~\cite{huatuogpto1}. Note that the base model Meditron-3-8B is not included since it's not an instruction-tuned model and can hardly follow evaluation instructions. We use default hyperparameters for baseline models, and all the thinking models are evaluated with thinking mode turned on. For each model, we adjust the tool-calling instruction and parsing logic to accept the supported tool-call format.

\paragraph{Evaluation environment.}
For evaluation, we use the same system prompts and action resolving logic as in RL (Section~\ref{sec:env_construction}). For patient simulator, during evaluation we use \texttt{gpt-4.1-mini}, whereas \texttt{gpt-4o-mini} is used for SFT data generation and \texttt{Qwen3-30B-A3B} for RL rollouts. This separation prevents overfitting to a specific simulator style. We evaluate all models on the original (unperturbed) test sets and we analyze the effect of noise injection separately via training-time ablations in Section~\ref{ablation_studies}. To ensure a fair comparison with baseline models, noise is not included in the test sets. \rev{We evaluate the robustness of our model under test-time noise separately in Section~\ref{sec:robustness}.}

\paragraph{Evaluation metrics.}
We employ an LLM-judge (\texttt{gpt-4.1-mini}) to evaluate diagnostic accuracy. The judge model first splits ground truth diagnosis and predicted diagnosis each into a set of individual medical conditions. Then, it judges the number of overlaps between the two sets. Given the judge's matched condition counts, we compute Jaccard similarity and strict diagnosis accuracy:
\begin{equation}\label{eq:jac_acc_eval}
\text{Jac} = \frac{|\mathcal{G} \cap \mathcal{P}|}{|\mathcal{G} \cup \mathcal{P}|}, \qquad \text{Acc} = \mathbbm{1}[\mathcal{G\subseteq\mathcal{P}}].
\end{equation}
where $\mathbbm{1}$ is the indicator function, $\mathcal{G}$ and $\mathcal{P}$ denote the sets of ground-truth and predicted conditions, respectively. The Jaccard score naturally handles multi-label diagnoses and reduces to binary scoring for single-condition cases. Since the diagnosis reward $R_{\text{dx}}$ (Equation~\ref{eq:r_dx}) used during RL also utilizes Jaccard score, there is a possibility of our trained model overfitting to this metric. Therefore, to ensure a fair comparison, we also measure strict diagnosis accuracy ($\text{Acc}$), where the model is required to at least predict all ground truth conditions to be considered correct. \rev{Since the same judge model is used for $R_{\text{dx}}$ during RL, we further verify our results with three other judge models in Appendix~\ref{sec:llm_judge_verify}.}
We also report embedding cosine similarity between predicted and ground-truth diagnoses:

\begin{equation}
  \text{Sim} = \max\left(0,\boldsymbol{\phi}(p)^\top\boldsymbol{\phi}(g)\right)
\end{equation}

where $\phi$ is MedEmbed-base-v0.1~\cite{balachandran2024medembed}, the embedding model we use.

\section{Results and Discussions}\label{sec:results_discussions}

\subsection{Performance Comparison}

\begin{table}[t]
\centering
\caption{Diagnostic performance across three benchmarks. \textbf{Sim} is cosine similarity between embeddings of the predicted and ground-truth diagnoses; \textbf{Jac} is the LLM-judge Jaccard similarity (matched concepts divided by the union of predicted and ground-truth concepts); \textbf{Acc} is LLM-judge strict accuracy (1 if every ground-truth concept was matched, 0 otherwise). \textbf{Bold} marks the best result per column, \underline{underline} marks the second best. Superscripts denote paired-bootstrap significance of MedExAgent-8B vs.\ the runner-up \rev{on each metric} ($B=10{,}000$; $^*p<0.05$, $^{**}p<0.01$, $^{***}p<0.001$; no marker = not significant at $p<0.05$).}
\label{tab:main_results}
\setlength{\tabcolsep}{2.5pt} 
\small
\begin{tabular}{@{}l lll lll lll@{}}
\toprule
& \multicolumn{3}{c}{\textbf{DDXPlus}}
& \multicolumn{3}{c}{\textbf{PMC-Patients}}
& \multicolumn{3}{c}{\textbf{AgentClinic (OOD)}} \\
\cmidrule(lr){2-4}\cmidrule(lr){5-7}\cmidrule(lr){8-10}
\textbf{Model} & Sim & Jac & Acc & Sim & Jac & Acc & Sim & Jac & Acc \\
\midrule
\multicolumn{10}{@{}l}{\emph{General open-source}} \\
Llama-3.1-8B-Instruct  & 0.640 & 0.383 & 0.605 & 0.507 & 0.122 & 0.157 & 0.575 & 0.225 & 0.298 \\
Qwen3-8B               & 0.605 & 0.328 & 0.460 & 0.532 & 0.092 & 0.120 & 0.565 & 0.153 & 0.219 \\
Qwen3-14B              & 0.659 & 0.399 & 0.538 & 0.565 & 0.113 & 0.143 & 0.614 & 0.226 & 0.284 \\
Qwen3-32B              & 0.685 & 0.466 & 0.626 & 0.551 & 0.127 & 0.157 & 0.637 & 0.310 & 0.368 \\
\midrule
\multicolumn{10}{@{}l}{\emph{Medical open-source}} \\
MedGemma-27B-text-it   & 0.498 & 0.241 & 0.325 & 0.469 & 0.108 & 0.123 & 0.544 & 0.214 & 0.249 \\
Aloe-Beta-70B          & \underline{0.716} & \rev{\underline{0.503}} & \rev{\underline{0.670}} & \underline{0.590} & \rev{\underline{0.171}} & \rev{\underline{0.219}} & \textbf{0.684} & \textbf{0.395} & \textbf{0.395} \\
HuatuoGPT-o1-70B       & 0.664 & 0.410 & 0.470 & 0.534 & 0.120 & 0.124 & 0.625 & 0.238 & 0.249 \\
\midrule
\textbf{MedExAgent-8B (Ours)} & \textbf{0.953}$^{***}$ & \textbf{0.937}$^{***}$ & \textbf{0.939}$^{***}$ & \textbf{0.626}$^{***}$ & \textbf{0.345}$^{***}$ & \textbf{0.314}$^{***}$ & \underline{0.672} & \underline{0.378} & \underline{0.378} \\
\bottomrule
\end{tabular}
\end{table}

\begin{table}[t]
  \centering
  \caption{Ablation study on training data, training stages, and reward design. Same metrics as Table~\ref{tab:main_results}. All rows are 8B variants from our pipeline; \textbf{bold} marks the best value per column, \underline{underline} marks the second best. Superscripts denote paired-bootstrap significance of MedExAgent-8B (Full) vs.\ the ablation variant on Sim ($B=10{,}000$; $^*p<0.05$, $^{**}p<0.01$, $^{***}p<0.001$).}
  \label{tab:ablations}
  \small
  \setlength{\tabcolsep}{2pt}
  \begin{tabular}{@{}l lll lll lll@{}}
  \toprule
  & \multicolumn{3}{c}{\textbf{DDXPlus}}
  & \multicolumn{3}{c}{\textbf{PMC-Patients}}
  & \multicolumn{3}{c}{\textbf{AgentClinic (OOD)}} \\
  \cmidrule(lr){2-4}\cmidrule(lr){5-7}\cmidrule(lr){8-10}
  \textbf{Model} & Sim & Jac & Acc & Sim & Jac & Acc & Sim & Jac & Acc \\
  \midrule
  \textbf{MedExAgent-8B}
    & 0.953 & 0.937 & 0.939
    & \textbf{0.626} & \textbf{0.345} & \textbf{0.314}
    & \textbf{0.672} & \textbf{0.378} & \textbf{0.378} \\
  \midrule
  \multicolumn{10}{@{}l}{\emph{Training data}} \\
  \quad w/o noise injection
    & \textbf{0.966} & \underline{0.943} & \underline{0.944}
    & \underline{0.624} & \underline{0.269} & \underline{0.254}
    & \underline{0.669} & \underline{0.318} & \underline{0.318} \\
  \midrule
  \multicolumn{10}{@{}l}{\emph{Training stages}} \\
  \quad SFT-only
    & 0.957 & \textbf{0.946} & \textbf{0.946}
    & 0.506$^{***}$ & 0.231 & 0.221
    & 0.589$^{***}$ & 0.286 & 0.289 \\
  \midrule
  \multicolumn{10}{@{}l}{\emph{Reward design}} \\
  \quad w/o tool-match reward ($w_{\text{tool}}{=}0$)
    & \underline{0.962} & 0.930 & 0.929
    & 0.593$^{***}$ & 0.219 & 0.203
    & 0.653 & 0.266 & 0.269 \\
  \quad w/o cost penalty ($w_{\text{cost}}{=}0$)
    & 0.846$^{***}$ & 0.831 & 0.831
    & 0.160$^{***}$ & 0.102 & 0.094
    & 0.383$^{***}$ & 0.261 & 0.264 \\
  \bottomrule
  \end{tabular}
\end{table}

\begin{table}[t]
  \centering
  \caption{Tool-use efficiency for the reward function ablation variants. \textbf{Calls} is the average number of tool calls per case (reported for context). \textbf{Call F1} is the harmonic mean of call-weighted precision and set-level recall over the ground-truth exam set; \textbf{\$ F1} is the harmonic mean of cost-weighted call-precision and cost-weighted recall. \rev{\textbf{Cost} is the average aggregate cost of exams per case on PMC-Patients-v2 (Section~\ref{sec:reward}).} \textbf{Bold} marks the column maximum.}
  \label{tab:ablation_tool_f1_summary}
  \setlength{\tabcolsep}{3pt}
  \small
  \begin{tabular}{@{}l ccc cccc ccc@{}}
  \toprule
  & \multicolumn{3}{c}{\textbf{DDxPlus}}
  & \multicolumn{4}{c}{\textbf{PMC-Patients-v2}}
  & \multicolumn{3}{c}{\textbf{AgentClinic (OOD)}} \\
  \cmidrule(lr){2-4}\cmidrule(lr){5-8}\cmidrule(lr){9-11}
  \textbf{Variant} & Calls & Call F1 & \$ F1 & Calls & Call F1 & \$ F1 & \rev{Cost} & Calls & Call F1 & \$ F1 \\
  \midrule
  \textbf{MedExAgent-8B}                                      & 3.75 & \textbf{0.828} & \textbf{0.810} & 4.03 & \textbf{0.701} & \textbf{0.705} & \rev{12.68} & 3.05 & \textbf{0.759} & \textbf{0.717} \\
  w/o tool-match reward ($w_{\text{tool}}{=}0$)& 2.38 & 0.768          & 0.755          & 1.99 & 0.592          & 0.581          & \rev{6.44} & 1.59 & 0.484          & 0.461          \\
  w/o cost penalty ($w_{\text{cost}}{=}0$)     & 5.65 & 0.741          & 0.711          & 9.51 & 0.648          & 0.659          & \rev{32.35} & 7.68 & 0.691          & 0.649          \\
  \bottomrule
  \end{tabular}
\end{table}

We first measure the performance of all compared models on the three test sets we used. We test the head-to-head significance of MedExAgent-8B against the open-source baselines on the same split via paired percentile bootstrap with $B=10{,}000$ resamples. Results are shown in Table~\ref{tab:main_results}.

As an in-distribution sanity check, MedExAgent-8B significantly outperforms every baseline on DDxPlus and PMC-Patients-v2 ($p<0.001$ on all metrics). The result of interest, however, is the held-out AgentClinic-MedQA benchmark, which evaluates all models on cases drawn from a distribution unseen during training: MedExAgent-8B is statistically tied with the 70B Aloe-Beta on Sim, Jac, and Acc; it also ties Qwen3-32B on strict accuracy ($p=0.857$) while outperforming the remaining five baselines on Jac. At 8B parameters, MedExAgent matches the strongest medical specialist in the comparison at roughly $9\times$ smaller parameter count. \rev{We analyze the failure cases of MedExAgent in Appendix~\ref{sec:error_analysis}.}

\subsection{Ablation Studies}
\label{ablation_studies}

\paragraph{Ablation of reward function terms.}
The RL reward (Equation~\ref{eq:reward-function}) decomposes into a diagnosis term, a tool-call term, and an exam-cost penalty.
To measure the contribution of each reward term, we train two otherwise identical variants with $w_\text{tool}=0$ and $w_{\text{cost}}=0$ each. Evaluation results are shown in Table~\ref{tab:ablations}.
We can observe that removing either $w_\text{tool}$ or $w_\text{cost}$ degrades performance.
As shown in Table~\ref{tab:ablation_tool_f1_summary}, when $w_{\text{cost}}$ is removed, even if $w_\text{tool}$ still places some constraints on excessive tool calls, the model still learns to call much more unnecessary exams, potentially due to a tendency to hack the diagnosis reward by gathering as much information as possible. In contrast, when $w_{\text{tool}}$ is removed, the model learns to call much less exams, potentially in response to the cost penalty $w_{\text{cost}}$. This demonstrates the necessity of including both $w_{\text{tool}}$ and $w_{\text{cost}}$ in our reward function. \rev{We also report the aggregate exam cost on PMC-Patients-v2 in Table~\ref{tab:ablation_tool_f1_summary}. When $w_{\text{cost}}$ is removed, the aggregate cost increases by $2.6\times$, since the model also orders more expensive exams. MedExAgent spends $1.11\times$ the cost of the recorded ground-truth exams and achieves the lowest cost per correct diagnosis among all baselines (Appendix~\ref{sec:cost_table}).}

\paragraph{Ablation of training-data noise.}
In Section~\ref{sec:noise_modeling} we introduced two synthetic noise channels: patient noise and exam noise, which are applied to both SFT and RL training stages.
To measure the contribution of noise injection, we train an otherwise identical variant with both noise channels disabled. Evaluation results are shown in Table~\ref{tab:ablations}.
We can observe that removing noise injection degrades Jac by \rev{$0.076$} on PMC-Patients-v2 and \rev{$0.060$} on AgentClinic, while leaving DDxPlus essentially unchanged ($\pm 0.005$ Jac, within run-to-run variation). 

\paragraph{Ablation of training stages.}
Our main pipeline finetunes the base model with 5K data entries for SFT followed by 20K entries for RL.
To measure the contribution of the RL stage, we train an otherwise identical variant on all 25K entries with SFT only and no RL. Evaluation results are shown in Table~\ref{tab:ablations}.
We can observe that removing RL degrades performance significantly on AgentClinic and PMC-Patients-v2, while matching the full model on the DDxPlus benchmark. This demonstrates that DDxPlus is an easier dataset that can be saturated by memorization in the SFT stage, while the more difficult in-distribution set PMC-Patients-v2 and the OOD set AgentClinic requires more generalization ability that SFT can not satisfy. 

{\revstart
\subsection{Robustness to Test-Time Noise}\label{sec:robustness}

\begin{table}[t]
  \centering
  \caption{Jac on PMC-Patients-v2 under false-negative noise with rate $\alpha$.}
  \label{tab:robustness}
  \small
  \setlength{\tabcolsep}{5pt}
  \begin{tabular}{@{}l cccccc@{}}
  \toprule
  $\alpha$ & 0 (clean) & 0.035 & 0.078 & 0.25 & 0.50 & 1.0 \\
  \midrule
  MedExAgent-8B & \textbf{0.345} & \textbf{0.334} & \textbf{0.339} & \textbf{0.313} & \textbf{0.271} & \textbf{0.202} \\
  w/o noise injection & 0.269 & 0.273 & 0.257 & 0.236 & 0.225 & 0.156 \\
  \bottomrule
  \end{tabular}
\end{table}

To evaluate the robustness of MedExAgent under test-time noise without leaking the noise types used in training, we use a different type of noise, false-negative, where the findings of each exam are removed with a probability of $\alpha$. The reported false-negative rate is 7.8\% for CT~\cite{alhammami2024assessment} and 17.7\% for chest X-ray~\cite{bradley2021chest}. Results on PMC-Patients-v2 are shown in Table~\ref{tab:robustness}. We can observe that the performance of MedExAgent degrades gradually as $\alpha$ increases, with a Jac drop of at most 0.032 when $\alpha\le0.25$. Additionally, the variant trained without noise injection performs worse than MedExAgent at every noise level, suggesting that training with noise helps the model avoid over-relying on individual exam results.
}

\section{Limitations and Future Work}

\paragraph{Assumption of patient invariance over time.} In our proposed environment and model ( Section~\ref{sec:problem_formulation}), we assume that the ground truth disease $d$ does not change over the entire episode. Additionally, we also assume that the observation function $O(\omega\mid d, a)$ does not change over time. In reality, this can be false especially for patients with critical conditions, whose symptoms and ground truth disease can progress over time. We leave it for future work to extend towards this direction.

\paragraph{Data and rollout generation with simulated patients.} In this work, we synthesized training data and RL rollout conversations with LLM-based simulated patients. Although we applied the persona from PatientSim~\cite{patientsim} to improve the authenticity of our simulated patients, there could still be distribution shifts between our training and testing data and real-world patients. \rev{Our evaluation also does not include clinician review of the conversations or diagnostic decisions, and relies on an LLM judge, which agrees well with three other judges (Appendix~\ref{sec:llm_judge_verify}) but may not fully reflect clinical correctness.} Validation with human patients remains necessary before deployment.

\rev{\paragraph{Exam-dominated diagnosis.} In our environment, removing all useful information from the patient only decreases Jac by 0.013 (Appendix~\ref{sec:channel_ablation}), while in real-world encounters, history taking provides most of the diagnostic information. This might be because the information that a doctor obtains during the conversation (e.g., physical examination) is covered by tool calls in our environment. We leave it for future work to design more informative conversations.}

\rev{\paragraph{Unrecorded exams.} In our environment, exams outside the recorded exam list return ``No significant findings'', so alternative exam strategies that are also clinically valid cannot be rewarded.}

\paragraph{English-only model.} Our proposed model is trained and tested with English datasets only. Clinical communications often involve culturally and linguistically specific descriptions, which can be hardly resolved by naively translating inputs and outputs to target languages. We leave it for future work to extend MedExAgent to multilingual clinical settings.
\section{Conclusion}

In this work, we addressed the gap between existing medical LLM benchmarks and the interactive, uncertain nature of real-world clinical diagnosis. We formalized clinical diagnosis as a POMDP with three action types: questioning the patient, ordering medical exams as tool calls, and issuing a diagnosis. We also introduced a systematic noise model comprising seven patient noise types and three exam noise types, together with diverse patient personas \cite{patientsim}. On top of this environment, we proposed MedExAgent, trained through a two-stage pipeline that combines supervised fine-tuning on Calgary-Cambridge-structured synthetic conversations with DAPO reinforcement learning against a composite reward capturing diagnostic accuracy, tool call quality, and examination cost.

Our experiments on DDxPlus, PMC-Patients-v2, and the out-of-distribution AgentClinic-MedQA benchmark show that MedExAgent-8B achieves diagnostic performance comparable to or exceeding significantly larger open-source baselines, while maintaining cost-efficient examination strategies. Ablation studies further confirm that both the tool call and exam cost reward terms, as well as noise injection during training, contribute meaningfully to the model's effectiveness and robustness.

By releasing our environment, evaluation pipeline, and model checkpoint, we hope to provide a realistic testbed and a strong open-source baseline for future research on interactive clinical diagnosis. Promising directions include extending the formulation to time-varying patient states and multi-visit episodes, calibrating the noise model against empirical clinical data, and validating diagnostic agents with real patients and physicians in the loop.

\clearpage

\bibliographystyle{abbrv}
\bibliography{bib}

\appendix

\section{Additional Experiment Details}

\subsection{Hyperparameters}\label{sec:hyperparameters}

In this section, we list the detailed hyperparameters used to finetune and evaluate our model in Table~\ref{tab:hyperparams}. For reward weights, we conducted a brief search over $(w_{\mathrm{tool}}, w_{\mathrm{cost}}) \in \{(1.0, 0.1), (0.5, 0.1), (0.1, 0.1), (0.5, 0.3)\}$ by running RL for 10 steps per configuration and selecting the configuration with the highest $R_{\mathrm{dx}}$ (the unweighted diagnosis reward). The configuration $(0.5, 0.1)$ was selected. 

The DAPO-specific hyperparameters in Table~\ref{tab:hyperparams} (group size, clip ratios, KL/entropy coefficients) are set to the defaults from~\cite{yu2025dapoopensourcellmreinforcement}. SFT hyperparameters follow standard values for instruction tuning at this scale. Batch sizes were set to the largest values that fit on our hardware. 

Noise injection probabilities $(p_{\mathrm{conv}}, p_{\mathrm{exam}}) = (0.3, 0.1)$ were chosen heuristically without tuning; the contribution of noise injection at these levels is validated by the ablation in Table~\ref{tab:ablations}.

\begin{table}[h]
  \centering
  \caption{Hyperparameters for SFT, RL, and evaluation.}
  \label{tab:hyperparams}
  \small
  \begin{tabular}{lll}
  \toprule
  \textbf{Hyperparameter} & \textbf{} & \textbf{Value} \\
  \midrule
  \multicolumn{3}{l}{\textit{Supervised Fine-Tuning}} \\
  \midrule
  Base model           &                           & OpenMeditron/Meditron3-8B \\
  Epochs               &                           & 1 \\
  Per-device batch size &                          & 2 \\
  Grad.\ accumulation  &                           & 2 \\
  Learning rate        &                           & $2{\times}10^{-5}$ \\
  LR schedule          &                           & cosine, 5\% warmup \\
  Optimizer            &                           & AdamW \\
  \midrule
  \multicolumn{3}{l}{\textit{Reinforcement Learning (DAPO)}} \\
  \midrule
  Group size $G$       &                           & 16 \\
  Rollout temperature  &                           & 1.0 \\
  Clip ratio (low, high) &                         & (0.2, 0.3) \\
  KL coefficient       &                           & 0.0 \\
  Entropy coefficient  &                           & 0.002 \\
  Learning rate        &                           & $1{\times}10^{-6}$ \\
  Global batch size    &                           & 64 \\
  Rollout noise        & patient / exam            & 0.3 / 0.1 \\
  Reward weights       & $(w_{\mathrm{tool}}, w_{\mathrm{cost}})$ & $(0.5,\,0.1)$ \\
  Diagnosis verifier model       &                           & \texttt{gpt-4.1-mini} \\
  Diagnosis verifier temperature          &                           & 0.7 \\
  \midrule
  \multicolumn{3}{l}{\textit{Evaluation}} \\
  \midrule
  Patient simulator model &                        & \texttt{gpt-4.1-mini} \\
  Patient simulator temperature &                  & 0.7 \\
  \bottomrule
  \end{tabular}
\end{table}

\newcommand{\prompt}[1]{{\ttfamily\small #1}}

\subsection{Compute Resources}\label{sec:compute_resources}

 \begin{table}[h]
    \centering
    \caption{Compute resource used to train MedExAgent.}
    \label{tab:compute}
    \small
    \begin{tabular}{ll}
    \toprule
    \textbf{Resource} & \textbf{Value} \\
    \midrule
    GPU model               & Nvidia RTX PRO 6000 Blackwell Max-Q \\
    Number of GPUs          & 4 \\
    VRAM per GPU            & 96\,GB \\
    CPU                     & AMD EPYC 9534, 128 cores \\
    System memory           & 1.5\,TiB \\
    CUDA / PyTorch          & 12.8 / 2.9 \\
    \bottomrule
    \end{tabular}
  \end{table}

The compute resource used to train MedExAgent is listed in Table~\ref{tab:compute}. With our settings and implementation, the SFT stage takes $\approx$ 4 hours and RL takes $\approx$ 5.5 days.

\clearpage
\section{Prompt Templates}\label{sec:prompt-templates}

\subsection{System Prompts}\label{sec:system_prompt_templates}

The system prompt templates applied in this paper are shown as follows. Patient prompt template is identical for both conversation generation and rollout generation during RL. The hidden canonical diagnosis in the doctor system prompt is hidden during rollout generation, and only kept for data generation.

\begin{promptbox1}[Data generation system prompt (Doctor)]
\ttfamily\small
You are an experienced physician conducting a medical consultation. Your goal is to gather information from the patient, order appropriate examinations, interpret results, and arrive at a diagnosis.

PATIENT DEMOGRAPHICS:
<demographics>

PATIENT'S MEDICAL HISTORY:
<medical\_history>

(\textbf{Only shown during data generation}) HIDDEN CANONICAL DIAGNOSIS (for training/consistency checking only; do NOT reveal to patient):
<hidden\_canonical\_diagnosis>

AVAILABLE EXAMINATIONS:
<tools\_block>

INSTRUCTIONS:

- Only output what you would say directly to the patient. Do NOT include your internal thoughts / reasoning / plan. 

- Pace gently, do NOT overwhelm the patient with too many questions at once. 

- Ask at most 1-2 short questions per turn.

- Order exams one at a time, do NOT order multiple exams in one turn.

- The hidden canonical diagnosis is ground truth for this case. Use it to guide your reasoning and keep the conversation clinically consistent, but do NOT reveal it to the patient before you have enough information and appropriate exam results.

- When you give the final diagnosis, it must match the hidden canonical diagnosis exactly.

- If no hidden canonical diagnosis is provided, reason from the conversation and exam results alone.

- You MUST end with [DIAGNOSIS: ...] to conclude the consultation.
\end{promptbox1}

\begin{promptbox2}[Data generation system prompt (Patient)]
\ttfamily\small
You are a patient visiting a doctor. You should role-play as a real patient would behave during a medical consultation.

YOUR DEMOGRAPHICS:
<demographics>

YOUR MEDICAL HISTORY:
<medical\_history>

YOUR CURRENT SYMPTOMS:
<self\_reported\_symptoms>

SELECTED PATIENT PERSONA:

Personality: [<p\_label>] <p\_instruction>

Language Proficiency: [<lang\_label>] <lang\_instruction>

Medical History Recall: [<recall\_label>] <recall\_instruction>

INSTRUCTIONS:

- Follow the selected persona naturally, but do NOT mention the persona labels to the doctor.

- Describe your symptoms using everyday, non-medical language when possible.

- Pace gently. Keep each reply brief and natural, usually 1-3 spoken sentences.

- Do NOT volunteer all your symptoms or history at once. Share information gradually as the doctor asks.

- Even if your selected personality is Verbose, do NOT dominate the conversation or dump long monologues unless the doctor explicitly asks for more detail.

- Only output spoken words. Do NOT include thoughts, reasoning, narration, stage directions, bracketed text, labels, or body language.

- Stay faithful to your true demographics and current symptoms. For past medical history, follow your selected recall setting exactly.
\end{promptbox2}

\subsection{Noise Injection}\label{sec:noise_injection_details}

As described in Section~\ref{sec:noise_modeling}, we inject seven types of patient noise and three types of exam noise during conversation generation for SFT and rollout generation for RL. Detailed definitions and examples for each noise type are shown in Table~\ref{tab:noise_types}. For each patient / exam, we first sample whether to introduce noise with designated noise levels (which are hyperparameters as shown in Table~\ref{tab:hyperparams}) as probability. If a patient is selected, we first decide 1-3 noise types according to eligibility, and assign each noise type to exactly one patient turn. If an exam is selected, we randomly assign one eligible noise type. 

In terms of noise injection method, for patient noise, we append a hint prompt after the patient prompt shown in Table~\ref{tab:noise_patterns}. For exam noise, we use a rule-based logic to postprocess the original exam results, as detailed in Table~\ref{tab:exam_noise}.

\subsection{LLM-Judge Prompt and Implementation}\label{sec:llm_judge_prompts}

We apply LLM-Judge to calculate both the diagnosis reward $R_\text{dx}$ during RL (Equation~\ref{eq:r_dx}) and the Jaccard score during evaluation (Equation~\ref{eq:jac_acc_eval}). The prompt is shown below.

\begin{promptbox1}[LLM Judge prompt]
\ttfamily\small
You are a medical expert evaluating a diagnosis prediction.

  Ground truth diagnosis: \{ground\_truth\}
  Predicted diagnosis: \{predicted\}

  Instructions:
  
  1. Identify the individual medical conditions in the ground truth. Note
     that a comma may be part of a single condition name (e.g.
     "seminoma, classic type" is ONE condition, "Follicular lymphoma,
     grade 2" is ONE condition). Semicolons or "and" typically separate
     distinct conditions.
     
  2. Identify the individual medical conditions in the prediction, using
     the same logic.
     
  3. For each ground truth condition, check if any predicted condition
     refers to the same disease. Consider synonyms (e.g. "heart attack" =
     "myocardial infarction"), abbreviations, and minor wording
     differences.
     
  4. Count how many ground truth conditions have a match in the
     predictions.

  You MUST respond in exactly this format (numbers only):
  
  gt\_count: <number of ground truth conditions>
  
  pred\_count: <number of predicted conditions>
  
  matched: <number of matched conditions>
\end{promptbox1}

Note that in this implementation, instead of directly asking for $|\mathcal{G}\cap\mathcal{P}|$ and $|\mathcal{G}\cup\mathcal{P}|$, we ask the LLM judge for $|\mathcal{G}|$, $|\mathcal{P}|$, and $|\mathcal{G}\cap\mathcal{P}|$, and calculate the Jaccard score and Accuracy with
\begin{equation}
    \text{Jac}=\frac{|\mathcal{G}\cap\mathcal{P}|}{|\mathcal{G}| + |\mathcal{P}| - |\mathcal{G}\cap\mathcal{P}|} \qquad
    \text{Acc}=\mathbbm{1}(|\mathcal{G}|=|\mathcal{G}\cap\mathcal{P}|)
\end{equation}
which is equivalent to Equation~\ref{eq:jac_acc_eval}. In this way we are able to calculate both metrics in one call to the LLM Judge.

\begin{table}[t]
  \centering
  \caption{Noise types applied to the observation function $O(\omega \mid d, a)$. Patient noises (white)
  affect conversational observations $\omega \in \Omega_{\text{conv}}$ and are applied with probability
  $p_{\text{conv}}$. Exam noises (gray) affect exam observations $\omega \in \Omega_{\text{exam}}$ and are
  applied with probability $p_{\text{exam}}$.}
  \label{tab:noise_types}
  \small
  \begin{tabular}{@{}llp{4.2cm}p{4.8cm}@{}}
  \toprule
  \textbf{Category} & \textbf{Noise Type} & \textbf{Description} & \textbf{Example} \\
  \midrule
  & Body Part Swap & Reports symptom location as an adjacent body part. & ``stomach pain'' $\rightarrow$ ``liver pain'' \\
  \cmidrule(l){2-4}
  & Symptom Confusion & Confuses a symptom descriptor with a related but incorrect term. & ``burning sensation'' $\rightarrow$ ``tingling sensation'' \\
  \cmidrule(l){2-4}
  & Severity Change & Misreports the intensity of symptoms, typically downplaying. & ``severe headache'' $\rightarrow$ ``moderate headache'' \\
  \cmidrule(l){2-4}
  \textbf{Patient} & Temporal Change & Misreports symptom duration or onset time. & ``3 weeks'' $\rightarrow$ ``1 week'' \\
  \cmidrule(l){2-4}
  & Omission & Fails to mention one symptom from a multi-symptom presentation. & Omits ``fatigue'' when also presenting with fever and cough. \\
  \cmidrule(l){2-4}
  & Self-Diagnosis & Volunteers an incorrect self-diagnosis based on lay sources. & ``I looked it up online and it seems like allergies.'' \\
  \cmidrule(l){2-4}
  & Vague Answer & Gives evasive, non-specific responses to direct questions. & ``I'm not really sure about that.'' \\
  \midrule
  & Body Part Swap & Replaces an anatomical location in findings with an adjacent body part. & ``Inflammation in the \textit{stomach} lining'' $\rightarrow$ ``Inflammation in the \textit{liver} lining'' \\
  \cmidrule(l){2-4}
  \textbf{Exam} & Omission & Drops one clause from multi-part findings. & ``Elevated WBC; low hemoglobin; normal platelets'' $\rightarrow$ ``Elevated WBC; normal platelets'' \\
  \cmidrule(l){2-4}
  & Ambiguity & Wraps findings in equivocal language indicating limited certainty. & ``Nodule in right lung'' $\rightarrow$ ``Findings are equivocal --- nodule in right lung. Cannot definitively rule out alternative interpretation.'' \\
  \bottomrule
  \end{tabular}
\end{table}

\begin{table}[t]
\centering
\caption{Hint templates used for injecting noise into patient responses. Prompts are emitted verbatim; placeholders are resolved as described.}
\label{tab:noise_patterns}
\small
\setlength{\tabcolsep}{4pt}
\renewcommand{\arraystretch}{1.1}
\begin{tabular}{@{}llp{8.5cm}@{}}
\toprule
\textbf{Category} & \textbf{Noise Type} & \textbf{Prompt (verbatim) and Placeholder Resolution} \\
\midrule

& Body Part Swap 
& \prompt{When answering, refer to your \{original\} symptom as being in your \{swapped\} area instead.}
\newline \textit{\{original\}}: first matched body-part keyword in symptoms; 
\textit{\{swapped\}}: sampled adjacent body part. \\
\cmidrule(l){2-3}

& Symptom Confusion 
& \prompt{When describing the sensation, say '\{confused\}' instead of '\{original\}'.} 
\newline \textit{\{original\}}: first matched descriptor; 
\textit{\{confused\}}: sampled confusion term. \\
\cmidrule(l){2-3}

& Severity Change 
& \prompt{When answering: \{severity\_instruction\}.} 
\newline Either: \textit{report the pain/symptom as X instead of Y} (if severity word matched), 
or \textit{downplay the severity of your symptoms slightly} (fallback). \\
\cmidrule(l){2-3}

\textbf{Patient} 
& Temporal Change 
& \prompt{When answering: \{temporal\_instruction\}.} 
\newline Either: \textit{say N units instead of M units} (if duration detected and perturbed), 
or \textit{be vague about when symptoms started — say `a while ago' or `recently'} (fallback). \\
\cmidrule(l){2-3}

& Omission 
& \prompt{In this response, do NOT mention \{omitted\_symptom\}. Talk only about your other symptoms or details.} 
\newline \textit{\{omitted\_symptom\}}: sampled from symptom list (requires $\geq$2 symptoms). \\
\cmidrule(l){2-3}

& Self-Diagnosis 
& \prompt{Work the following into your response naturally: '\{phrase\}' Then continue answering the doctor's question.} 
\newline \textit{\{phrase\}}: template filled with a sampled common condition (e.g., cold, flu, allergies). \\
\cmidrule(l){2-3}

& Vague Answer 
& \prompt{For this response, be vague and evasive. Say something like '\{vague\}' Do not provide specific details for this question.} 
\newline \textit{\{vague\}}: sampled from a fixed pool of non-committal responses. \\
\bottomrule
\end{tabular}
\end{table}

\begin{table}[t]

\centering

\caption{Exam noise transformations applied to tool-returned findings. Unlike patient noise, these are not prompts; transformations are applied directly to the raw findings string before delivery.}

\label{tab:exam_noise}

\small

\setlength{\tabcolsep}{4pt}

\renewcommand{\arraystretch}{1.1}

\begin{tabular}{@{}llp{8.5cm}@{}}

\toprule

\textbf{Category} & \textbf{Noise Type} & \textbf{Transformation (verbatim) and Resolution} \\

\midrule

& Body Part Swap 

& Replace the first body-part token with a randomly sampled adjacent term from a predefined body part list. \\

\cmidrule(l){2-3}

\textbf{Exam} 

& Omission 

& If multiple findings exist, drop a random non-empty one. \\

\cmidrule(l){2-3}

& Ambiguity 

& Wrap the original findings using one of the following templates (uniformly sampled): 

\begin{itemize}
\item \prompt{Findings are equivocal — \{original\}. Cannot definitively rule out alternative interpretation.}
\item \prompt{Results show \{original\}. However, findings are not entirely clear and may warrant further evaluation.}
\item \prompt{\{original\}. Note: image quality/sample quality limits definitive interpretation.} 
\end{itemize} \\

\bottomrule

\end{tabular}

\end{table}

\section{Verification of LLM-judge accuracy}\label{sec:llm_judge_verify}

The diagnosis reward $R_{\text{dx}}$ in Eq.~\eqref{eq:r_dx} relies on an LLM
judge (\texttt{gpt-4.1-mini}) to extract condition sets from the predicted
and ground-truth diagnoses, semantically match equivalent conditions, and
return the integer counts $|{\mathcal{G}}|$,
$|{\mathcal{P}}|$, $|{\mathcal{G}} \cap {\mathcal{P}}|$
from which the Jaccard score is computed (more details in Appendix~\ref{sec:llm_judge_prompts}). In this section we measure the judge's accuracy directly on conditions drawn from our test sets.

We sample 99 condition pairs from the
\href{https://disease-ontology.org}{Disease Ontology}~\cite{baron2026dokb}, restricted to DOID entries that match a diagnosis
appearing in DDxPlus, PMC-Patients-v2, or AgentClinic-MedQA. From the matched terms we deterministically draw three uniform buckets with 33 samples each:

\begin{itemize}
\item \textbf{Synonym}: Ground truth is test-set diagnosis; prediction is a DOID \texttt{EXACT} synonym. Expected $R_{\text{dx}}{=}1$.
\item \textbf{Distractor}: Ground truth is test-set diagnosis; prediction is a DOID sibling under the \texttt{is\_a} graph, with three coupling filters -- substring-disjoint names, different acuity prefixes (\{acute, chronic, recurrent, severe, mild\}), different trailing integers. Expected $R_{\text{dx}}{=}0$.
\item \textbf{Multi-condition partial}: Ground truth is ``$A$ and $B$'' for two
test-set-grounded terms with no shared parent and no \texttt{is\_a}
relation; prediction is $A$ alone. Expected $|{\mathcal{G}}|{=}2$,
$|{\mathcal{P}}|{=}1$, $|{\mathcal{G}} \cap {\mathcal{P}}|{=}1$,
$R_{\text{dx}}{=}\tfrac{1}{2}$.
\end{itemize}

Each pair is scored independently by \texttt{gpt-4.1-mini} at
temperature $0$ using the prompt template of
Appendix~\ref{sec:prompt-templates}. We parse the response
$\{\texttt{gt\_count}, \texttt{pred\_count}, \texttt{matched}\}$ and
compute $\widehat{R}_{\text{dx}}$ as in Equation~\ref{eq:r_dx}.

\begin{table}[h]
\centering
\caption{Per-bucket accuracy on the 99-pair probe (\texttt{gpt-4.1-mini},
temperature $0$). Accuracy is the ratio of the judge giving the correct answer. MAE is the mean absolute error between $\widehat{R}_{\text{dx}}$ and the expected score.}
\label{tab:judge_probe}
\small
\begin{tabular}{@{}lrrr@{}}
\toprule
\textbf{Bucket} & $n$ & \textbf{Accuracy} & \textbf{MAE} \\
\midrule
Synonym (matched as equivalent)            & 33 & 91\,\% & 0.197 \\
Distractor (correctly rejected)            & 33 & 94\,\% & 0.152 \\
Multi-condition partial & 33 & 100\,\% & 0.000 \\
\bottomrule
\end{tabular}
\end{table}

The judge is exact on multi-condition counting and partial matching. Synonym sensitivity is $30/33 = 91\%$ and distractor specificity is $31/33 = 94\%$; the remaining failures are mostly obscure historical eponyms (e.g.\ ``Rose's tamponade''; ``insular sclerosis'').

{\revstart
\paragraph{Agreement with other judges.} To further verify the judge on the actual evaluation workload, we collect 2,992 (ground truth, prediction) pairs from our evaluation runs, including MedExAgent and the no-noise ablation on PMC-Patients-v2 and AgentClinic-MedQA, and Aloe-Beta-8B and Qwen3-32B on PMC-Patients-v2. We re-score each pair with three recent models using the same judge prompt. Results are shown in Table~\ref{tab:judge_agreement}. We can observe that \texttt{gpt-4.1-mini} has high agreement with all three models.

\begin{table}[!htbp]
\centering
\caption{Agreement between \texttt{gpt-4.1-mini} and three other judge models on 2,992 evaluation pairs.}
\label{tab:judge_agreement}
\small
\begin{tabular}{@{}lcc@{}}
\toprule
\texttt{gpt-4.1-mini} vs. & Agreement & Cohen's $\kappa$ \\
\midrule
\texttt{gpt-5.5} & 93.98\% & 0.834 \\
\texttt{claude-sonnet-5} & 94.22\% & 0.844 \\
\texttt{qwen3.6-35b-a3b} & 96.02\% & 0.899 \\
\bottomrule
\end{tabular}
\end{table}

\paragraph{Evaluation with other judges.} Since \texttt{gpt-4.1-mini} is used for both the RL reward and evaluation, we further evaluate selected models with the three judges above. Results are shown in Table~\ref{tab:judge_swap}. We can observe that the four judges differ mainly in how permissive they are. The fraction of pairs judged as a match is 0.243 for \texttt{gpt-5.5}, 0.257 for \texttt{claude-sonnet-5}, 0.345 for \texttt{gpt-4.1-mini}, and 0.365 for \texttt{qwen3.6-35b-a3b}, so our training judge is not the most permissive one. MedExAgent outperforms Aloe-Beta-8B and Qwen3-32B on PMC-Patients-v2 under all four judges. On AgentClinic, the variant without noise injection scores higher than MedExAgent under \texttt{gpt-5.5} and \texttt{claude-sonnet-5}.

\begin{table}[!htbp]
\centering
\caption{Jac of selected models under four judge models, using the same judge prompt.}
\label{tab:judge_swap}
\small
\begin{tabular}{@{}lcccc@{}}
\toprule
System / test set & \texttt{gpt-4.1-mini} & \texttt{gpt-5.5} & \texttt{claude-sonnet-5} & \texttt{qwen3.6} \\
\midrule
MedExAgent-8B, PMC & 0.337 & 0.248 & 0.267 & 0.359 \\
w/o noise injection, PMC & 0.253 & 0.219 & 0.226 & 0.259 \\
Aloe-Beta-8B, PMC & 0.114 & 0.110 & 0.118 & 0.132 \\
Qwen3-32B, PMC & 0.110 & 0.088 & 0.131 & 0.125 \\
\midrule
MedExAgent-8B, AgentClinic & 0.359 & 0.261 & 0.281 & 0.407 \\
w/o noise injection, AgentClinic & 0.358 & 0.309 & 0.313 & 0.358 \\
\bottomrule
\end{tabular}
\end{table}

\section{Additional Analyses}\label{sec:additional_analyses}

\subsection{Ablation of Conversation and Exam Stages}\label{sec:channel_ablation}
To analyze how much the conversation and the exams contribute to the diagnosis in our environment, we ablate each of them by removing all useful information from that stage. Results are shown in Table~\ref{tab:channel_ablation}. We can observe that removing the useful information from the patient only decreases Jac by 0.013, while making every ground-truth exam return a negative result decreases Jac by 0.138. Since the exam results are taken from the source cases rather than generated by LLMs, this shows that the effect of patient simulation on our results is limited.

\begin{table}[!htbp]
\centering
\caption{Ablation of the conversation and exam stages for MedExAgent-8B on PMC-Patients-v2. Rows below the first show the change from the clean setting.}
\label{tab:channel_ablation}
\small
\begin{tabular}{@{}lccc@{}}
\toprule
Condition & Jac & Acc & Sim \\
\midrule
Clean & 0.339 & 0.302 & 0.625 \\
Patient withholds every symptom, location, and timing detail & $-0.013$ & $-0.003$ & $-0.002$ \\
Every ground-truth exam returns a negative & $-0.138$ & $-0.113$ & $-0.045$ \\
Both & $-0.141$ & $-0.118$ & $-0.055$ \\
\bottomrule
\end{tabular}
\end{table}

\subsection{Effect of Patient Personas}\label{sec:persona_ablation}
We first verify that the persona changes the behavior of the simulated patient as intended. Across 10,026 conversations, the average number of words per patient turn ranges from 78.8 (verbose) to 24.3 (plain), and the disclosure rate of medical history is 0.349, 0.243, and 0.130 for high, low, and no recall, respectively. We then remove the persona from the patient prompt and re-run MedExAgent and Qwen3-8B on the same PMC-Patients-v2 cases, keeping all other settings fixed. Results are shown in Table~\ref{tab:persona_ablation}. We can observe that neither model has a significant persona effect, and the difference between their effects is also not significant (Jac $p=0.360$, Acc $p=0.788$). Accuracy broken down by personality, language proficiency, and history recall also shows no significant difference for MedExAgent or three baselines on the same 1,000 cases (all $p\ge0.224$). Therefore, the personas produce the intended differences in patient communication, but do not significantly affect the diagnostic difficulty in our environment.

\begin{table}[!htbp]
\centering
\caption{Performance with and without patient personas on PMC-Patients-v2, with paired $p$-values.}
\label{tab:persona_ablation}
\small
\begin{tabular}{@{}lrcc@{}}
\toprule
Model & $n$ & Jac, on $\rightarrow$ off ($\Delta$, $p$) & Acc, on $\rightarrow$ off ($\Delta$, $p$) \\
\midrule
MedExAgent-8B & 1,000 & 0.3393 $\rightarrow$ 0.3446 ($+0.0053$, 0.622) & 0.3020 $\rightarrow$ 0.3190 ($+0.0170$, 0.130) \\
Qwen3-8B & 990 & 0.0895 $\rightarrow$ 0.1076 ($+0.0181$, 0.055) & 0.1152 $\rightarrow$ 0.1374 ($+0.0222$, 0.066) \\
\bottomrule
\end{tabular}
\end{table}

\subsection{Conversation Generation without Ground Truth Diagnosis}\label{sec:hidden_dx_ablation}
During SFT data generation, the hidden canonical diagnosis is given in the doctor prompt (Appendix~\ref{sec:prompt-templates}) to make sure that the generated conversation does not diverge from the ground truth diagnosis. Using a known target to generate the path leading to it is a common practice in synthetic supervision. For example, STaR~\cite{zelikman2022star} gives the correct answer to the model when generating rationales for failed problems. To test whether the diagnosis is needed, we sample 60 PMC-Patients-v2 cases with a fixed random seed and generate their conversations twice, changing only whether the doctor model sees the diagnosis. Results are shown in Table~\ref{tab:hidden_dx}. We can observe that without the diagnosis, 40\% of the conversations end with a medically different diagnosis, and only 3.3\% match the source diagnosis closely enough to pass the filter in our generation pipeline. Training on these conversations would introduce contradictory supervision.

\begin{table}[!htbp]
\centering
\caption{SFT conversation generation with and without the ground truth diagnosis in the doctor prompt, on 60 PMC-Patients-v2 cases.}
\label{tab:hidden_dx}
\small
\begin{tabular}{@{}lcc@{}}
\toprule
 & With diagnosis & Without diagnosis \\
\midrule
Final diagnosis matches exactly & 98.3\% & 3.3\% \\
Final diagnosis correct under LLM judge & 100\% & 60\% \\
Judge Jaccard & 1.000 & 0.560 \\
\bottomrule
\end{tabular}
\end{table}

\subsection{Aggregate Exam Cost}\label{sec:cost_table}
We compare the aggregate exam cost of MedExAgent with the baseline models and the recorded ground-truth exams on PMC-Patients-v2. Cost per correct diagnosis is calculated as aggregate cost divided by strict accuracy. Results are shown in Table~\ref{tab:cost_baselines}. We can observe that MedExAgent spends approximately the same cost as the recorded ground-truth exams ($1.11\times$), and achieves the lowest cost per correct diagnosis among all baseline models.

\begin{table}[!htbp]
\centering
\caption{Aggregate exam cost of all models on PMC-Patients-v2.}
\label{tab:cost_baselines}
\small
\setlength{\tabcolsep}{4pt}
\begin{tabular}{@{}lcccccc@{}}
\toprule
System & Calls & Agg.\ cost & $\times$ GT workup & Jac & Acc & Cost / correct dx \\
\midrule
Recorded ground-truth workup & 3.46 & 11.46 & 1.00 & -- & -- & -- \\
\midrule
Llama-3.1-8B-Instruct & 6.59 & 19.15 & 1.67 & 0.122 & 0.157 & 122.0 \\
MedGemma-27B-text-it & 5.84 & 17.05 & 1.49 & 0.109 & 0.123 & 138.7 \\
Aloe-Beta-70B & 4.60 & 14.42 & 1.26 & 0.171 & 0.219 & 65.8 \\
Qwen3-32B & 4.66 & 14.05 & 1.23 & 0.127 & 0.157 & 89.5 \\
HuatuoGPT-o1-70B & 3.97 & 12.50 & 1.09 & 0.120 & 0.124 & 100.6 \\
Qwen3-14B & 3.14 & 9.45 & 0.82 & 0.113 & 0.143 & 66.1 \\
Qwen3-8B & 3.07 & 8.76 & 0.76 & 0.092 & 0.120 & 73.0 \\
\midrule
w/o cost penalty ($w_{\text{cost}}{=}0$) & 9.51 & 32.35 & 2.82 & 0.102 & 0.094 & 344.1 \\
w/o noise injection & 3.44 & 10.53 & 0.92 & 0.269 & 0.254 & 41.5 \\
w/o tool-match ($w_{\text{tool}}{=}0$) & 1.99 & 6.44 & 0.56 & 0.219 & 0.203 & 31.7 \\
MedExAgent-8B (ours) & 4.03 & 12.68 & 1.11 & 0.345 & 0.314 & 40.4 \\
\bottomrule
\end{tabular}
\end{table}

\section{Error Analysis}\label{sec:error_analysis}
\paragraph{Failure stage.} We first analyze at which stage of the conversation the failures happen. Specifically, we prompt \texttt{gpt-5.5}, \texttt{claude-sonnet-5}, and \texttt{gpt-4.1-mini} to classify which stage resulted in the final error, and report the majority vote. Results are shown in Table~\ref{tab:error_stage}. For PMC-Patients-v2 and AgentClinic, most errors are due to exam selection, which is the most impactful and difficult step. For DDxPlus, where the cases are relatively easy, final reasoning becomes the main reason for errors.

\begin{table}[!htbp]
\centering
\caption{Primary failure stage of MedExAgent-8B errors (majority vote of three judges).}
\label{tab:error_stage}
\small
\begin{tabular}{@{}lccc@{}}
\toprule
Primary failure stage & PMC ($n{=}692$) & DDxPlus ($n{=}50$) & AgentClinic ($n{=}125$) \\
\midrule
Exam selection & 50.7\% & 26.0\% & 52.8\% \\
Final reasoning & 29.3\% & 52.0\% & 28.8\% \\
Exam interpretation & 13.3\% & 6.0\% & 12.8\% \\
Questioning & 5.8\% & 16.0\% & 5.6\% \\
Label or evidence inconsistency & 0.7\% & 0.0\% & 0.0\% \\
\bottomrule
\end{tabular}
\end{table}

\paragraph{Failure modes.} To analyze the detailed reasons for failures, we sample 100 failure cases across the three datasets and manually read them to form a taxonomy of failure modes. We then prompt \texttt{gpt-5.5} to classify all the failure cases into these categories. Results are shown in Table~\ref{tab:error_modes}. Insufficient examination is the most common failure mode, which is consistent with our stage analysis. We further map our taxonomy into the diagnostic process stages in Schiff et al.~\cite{schiff2009diagnostic}, which studies 583 physician-reported diagnostic errors (Table~\ref{tab:error_physician}). We can observe that the error distribution of MedExAgent is similar to that of human physicians, where testing is the most common source of errors.

\begin{table}[!htbp]
\centering
\caption{Failure modes of MedExAgent-8B classified by \texttt{gpt-5.5}.}
\label{tab:error_modes}
\small
\begin{tabular}{@{}lcccc@{}}
\toprule
Failure mode & PMC & DDxPlus & AgentClinic & Overall \\
\midrule
Insufficient examination & 26.9\% & 10.7\% & 39.2\% & 27.6\% \\
Incidental anchoring & 16.9\% & 30.4\% & 12.0\% & 17.1\% \\
Pattern synthesis & 10.8\% & 30.4\% & 19.2\% & 13.3\% \\
Evidence misread & 15.2\% & 8.9\% & 10.4\% & 14.1\% \\
Negative-evidence mishandling & 11.0\% & 0.0\% & 6.4\% & 9.6\% \\
History omission & 7.4\% & 14.3\% & 8.0\% & 7.9\% \\
Specificity failure & 6.2\% & 1.8\% & 4.0\% & 5.6\% \\
Other & 5.6\% & 3.6\% & 0.8\% & 4.8\% \\
\bottomrule
\end{tabular}
\end{table}

\begin{table}[!htbp]
\centering
\caption{Errors of MedExAgent-8B and human physicians~\cite{schiff2009diagnostic} by diagnostic process stage.}
\label{tab:error_physician}
\small
\begin{tabular}{@{}lcc@{}}
\toprule
Diagnostic process stage & MedExAgent ($n{=}867$) & Physicians~\cite{schiff2009diagnostic} ($n{=}583$) \\
\midrule
Testing: ordering, processing, acting on results & 43.9\% & 44\% \\
Clinician assessment & 42.5\% & 32\% \\
History taking & 4.6\% & 10\% \\
Physical examination & 7.4\% & 10\% \\
Other & 1.6\% & 3\% \\
\bottomrule
\end{tabular}
\end{table}
}

\clearpage
\section{Additional Experiment Result Tables}\label{sec:add_exp_results}

Table~\ref{tab:cis_all} reports 95\% percentile-bootstrap confidence intervals (B = 10,000) for all models and ablation variants across the three benchmarks.

\begin{table}[t]
  \centering
  \caption{95\% confidence intervals across datasets. Each cell reports mean\,$\pm$\,half-width of the 95\% percentile-bootstrap CI over per-sample scorer outputs ($B=10{,}000$). Same metrics, bolding, and variants as Tables~\ref{tab:main_results} and~\ref{tab:ablations}. \textbf{Bold} marks the column-best among non-ablation rows, \underline{underline} marks the second best; ablation rows are listed for reference.}
  \label{tab:cis_all}
  \small
  \setlength{\tabcolsep}{4pt}

  \begin{subtable}[t]{\linewidth}
  \centering
  \caption{AgentClinic (OOD, $N{=}201$)}
  \begin{tabular}{@{}lccc@{}}
  \toprule
  \textbf{Model / Variant} & Sim & Jac & Acc \\
  \midrule
  Llama-3.1-8B-Instruct & 0.575$\,\pm\,$0.033 & 0.225$\,\pm\,$0.053 & 0.298$\,\pm\,$0.065 \\
  Qwen3-8B              & 0.565$\,\pm\,$0.017 & 0.153$\,\pm\,$0.044 & 0.219$\,\pm\,$0.057 \\
  Qwen3-14B             & 0.614$\,\pm\,$0.022 & 0.226$\,\pm\,$0.053 & 0.284$\,\pm\,$0.062 \\
  Qwen3-32B             & 0.637$\,\pm\,$0.028 & 0.310$\,\pm\,$0.061 & 0.368$\,\pm\,$0.067 \\
  MedGemma-27B-text-it  & 0.544$\,\pm\,$0.038 & 0.214$\,\pm\,$0.053 & 0.249$\,\pm\,$0.060 \\
  Aloe-Beta-70B         & \textbf{0.684$\,\pm\,$0.023} & \textbf{0.395$\,\pm\,$0.061} & \textbf{0.395$\,\pm\,$0.061} \\
  HuatuoGPT-o1-70B      & 0.625$\,\pm\,$0.025 & 0.238$\,\pm\,$0.059 & 0.249$\,\pm\,$0.062 \\
  \textbf{MedExAgent-8B (full)} & \underline{0.672$\,\pm\,$0.024} & \underline{0.378$\,\pm\,$0.067} & \underline{0.378$\,\pm\,$0.067} \\
  \midrule
  \quad w/o noise injection     & 0.669$\,\pm\,$0.024 & 0.318$\,\pm\,$0.065 & 0.318$\,\pm\,$0.065 \\
  \quad SFT-only                & 0.589$\,\pm\,$0.038 & 0.286$\,\pm\,$0.063 & 0.289$\,\pm\,$0.063 \\
  \quad w/o tool-match reward   & 0.653$\,\pm\,$0.026 & 0.266$\,\pm\,$0.061 & 0.269$\,\pm\,$0.061 \\
  \quad w/o cost penalty        & 0.383$\,\pm\,$0.055 & 0.261$\,\pm\,$0.062 & 0.264$\,\pm\,$0.062 \\
  \bottomrule
  \end{tabular}
  \end{subtable}

  \begin{subtable}[t]{\linewidth}
  \centering
  \caption{DDxPlus ($N{\approx}1000$)}
  \begin{tabular}{@{}lccc@{}}
  \toprule
  \textbf{Model / Variant} & Sim & Jac & Acc \\
  \midrule
  Llama-3.1-8B-Instruct & 0.640$\,\pm\,$0.013 & 0.383$\,\pm\,$0.023 & 0.605$\,\pm\,$0.030 \\
  Qwen3-8B              & 0.605$\,\pm\,$0.009 & 0.328$\,\pm\,$0.024 & 0.460$\,\pm\,$0.030 \\
  Qwen3-14B             & 0.659$\,\pm\,$0.009 & 0.399$\,\pm\,$0.026 & 0.538$\,\pm\,$0.032 \\
  Qwen3-32B             & 0.685$\,\pm\,$0.012 & 0.466$\,\pm\,$0.025 & 0.626$\,\pm\,$0.030 \\
  MedGemma-27B-text-it  & 0.498$\,\pm\,$0.019 & 0.241$\,\pm\,$0.023 & 0.325$\,\pm\,$0.029 \\
  Aloe-Beta-70B         & \underline{0.716$\,\pm\,$0.009} & \underline{0.503$\,\pm\,$0.025} & \underline{0.670$\,\pm\,$0.029} \\
  HuatuoGPT-o1-70B      & 0.664$\,\pm\,$0.013 & 0.410$\,\pm\,$0.027 & 0.470$\,\pm\,$0.031 \\
  \textbf{MedExAgent-8B (full)} & \textbf{0.953$\,\pm\,$0.008} & \textbf{0.937$\,\pm\,$0.015} & \textbf{0.939$\,\pm\,$0.015} \\
  \midrule
  \quad w/o noise injection     & 0.966$\,\pm\,$0.007 & 0.943$\,\pm\,$0.014 & 0.944$\,\pm\,$0.014 \\
  \quad SFT-only                & 0.957$\,\pm\,$0.010 & 0.946$\,\pm\,$0.013 & 0.946$\,\pm\,$0.013 \\
  \quad w/o tool-match reward   & 0.962$\,\pm\,$0.007 & 0.930$\,\pm\,$0.016 & 0.929$\,\pm\,$0.016 \\
  \quad w/o cost penalty        & 0.846$\,\pm\,$0.021 & 0.831$\,\pm\,$0.023 & 0.831$\,\pm\,$0.023 \\
  \bottomrule
  \end{tabular}
  \end{subtable}

  \begin{subtable}[t]{\linewidth}
  \centering
  \caption{PMC-Patients ($N{\approx}1000$)}
  \begin{tabular}{@{}lccc@{}}
  \toprule
  \textbf{Model / Variant} & Sim & Jac & Acc \\
  \midrule
  Llama-3.1-8B-Instruct & 0.507$\,\pm\,$0.013 & 0.122$\,\pm\,$0.018 & 0.157$\,\pm\,$0.022 \\
  Qwen3-8B              & 0.532$\,\pm\,$0.008 & 0.092$\,\pm\,$0.016 & 0.120$\,\pm\,$0.021 \\
  Qwen3-14B             & 0.565$\,\pm\,$0.007 & 0.113$\,\pm\,$0.017 & 0.143$\,\pm\,$0.022 \\
  Qwen3-32B             & 0.551$\,\pm\,$0.010 & 0.127$\,\pm\,$0.018 & 0.157$\,\pm\,$0.022 \\
  MedGemma-27B-text-it  & 0.469$\,\pm\,$0.014 & 0.108$\,\pm\,$0.017 & 0.123$\,\pm\,$0.021 \\
  Aloe-Beta-70B         & \underline{0.590$\,\pm\,$0.008} & \underline{0.171$\,\pm\,$0.021} & \underline{0.219$\,\pm\,$0.026} \\
  HuatuoGPT-o1-70B      & 0.534$\,\pm\,$0.010 & 0.120$\,\pm\,$0.019 & 0.124$\,\pm\,$0.021 \\
  \textbf{MedExAgent-8B (full)} & \textbf{0.626$\,\pm\,$0.010} & \textbf{0.344$\,\pm\,$0.028} & \textbf{0.314$\,\pm\,$0.029} \\
  \midrule
  \quad w/o noise injection     & 0.624$\,\pm\,$0.010 & 0.269$\,\pm\,$0.026 & 0.254$\,\pm\,$0.027 \\
  \quad SFT-only                & 0.506$\,\pm\,$0.018 & 0.231$\,\pm\,$0.026 & 0.221$\,\pm\,$0.026 \\
  \quad w/o tool-match reward   & 0.593$\,\pm\,$0.010 & 0.219$\,\pm\,$0.025 & 0.203$\,\pm\,$0.025 \\
  \quad w/o cost penalty        & 0.160$\,\pm\,$0.019 & 0.102$\,\pm\,$0.018 & 0.094$\,\pm\,$0.018 \\
  \bottomrule
  \end{tabular}
  \end{subtable}
\end{table}

\section{Additional Statements}\label{sec:disclaimers}

\subsection{Broader Impact Statement}\label{sec:broader_impact}

MedExAgent is a research prototype of an LLM-based agent for interactive clinical diagnosis. Our goal is to lower the compute barrier for medical agent research and provide an open-source baseline for on-premises deployment in privacy-sensitive research settings. We note that MedExAgent is not intended for clinical use. It is trained on synthetic conversations with LLM-simulated patients, and has not been validated in the real world. To prevent misuse, we gate access to the released checkpoint behind a usage agreement that restricts use to research and prohibits patient-facing deployment, and we include a prototype-only disclaimer in the paper, model card, and repository README.

\subsection{Assets Statement}\label{sec:released_assets}

\textbf{New assets.} We release three artifacts: (i) the MedExAgent-8B model checkpoint, (ii) the dataset used to train MedExAgent, including synthesized conversations for SFT and extracted data tables for RL, and (iii) the code used for data generation, model finetuning, and evaluation. 

MedExAgent-8B is finetuned from Meditron3-8B, which itself is finetuned from Llama-3.1-8B. Therefore, we release the checkpoint licensed under the Llama 3.1 Community License, with access gated behind a usage agreement that restricts use to research and prohibits patient-facing deployment. Our dataset is partially derived from PMC-Patients-v2~\cite{zhao2023large}, which is licensed under CC BY-NC-SA 4.0. Therefore, we release the dataset licensed under CC BY-NC-SA 4.0. The code is released under the MIT license.

\textbf{Existing assets.} Table~\ref{tab:asset_licenses} lists all existing assets used in this work, along with their licenses. All assets are used in accordance with their original terms of use. For PatientSim, we only use the patient persona published in the paper, which is licensed under CC BY-NC-ND 4.0 per arXiv. For CMS PFS RVU and CMS CLFS values, we do not redistribute raw values in any of our released assets. 

\begin{table}[h]
\centering
\caption{Licenses and sources for existing assets used in this work.}
\label{tab:asset_licenses}
\small
\begin{tabular}{@{}ll@{}}
\toprule
Asset & License \\
\midrule
DDxPlus~\cite{fansi2022ddxplus} & CC BY 4.0 \\
PMC-Patients-v2~\cite{zhao2023large} & CC BY-NC-SA 4.0 \\
AgentClinic-MedQA~\cite{schmidgall2024agentclinic} & MIT \\
PatientSim personas~\cite{patientsim} & CC BY-NC-ND 4.0 \\
Disease Ontology~\cite{baron2026dokb} & CC0 1.0 \\
Meditron3-8B~\cite{meditron} & Llama 3.1 Community License \\
Llama-3.1-8B-Instruct~\cite{grattafiori2024llama} & Llama 3.1 Community License \\
Qwen3-8B/14B/32B~\cite{yang2025qwen3} & Apache 2.0 \\
MedGemma-27B-text-it~\cite{sellergren2025medgemma} & Health AI Developer Foundations License \\
Aloe-Beta-70B~\cite{garcia2025aloe} & CC BY-NC 4.0 \\
HuatuoGPT-o1-70B~\cite{huatuogpto1} & Apache 2.0 \\
MedEmbed-base-v0.1~\cite{balachandran2024medembed} & Apache 2.0 \\
CMS PFS RVU 2026~\cite{cms_pfs_rvu26a} & U.S. government work; CPT content \copyright~AMA \\
CMS CLFS 2025Q4~\cite{cms_clfs_2025q4} & U.S. government work; CPT content \copyright~AMA \\
\bottomrule
\end{tabular}
\end{table}

\subsection{LLM Usage Statement}\label{sec:llm_usage}

LLMs are a core component of this work and are used in four distinct roles, each documented in the body of the paper:

\begin{itemize}
    \item \textbf{The trained agent.} MedExAgent itself is an LLM (Meditron3-8B fine-tuned via SFT and DAPO).
    \item \textbf{Synthetic conversation generation for SFT.} Section~\ref{sec:conversation_generation} and Appendix~\ref{sec:prompt-templates} document the use of \texttt{gpt-4o-mini} as both the doctor and patient simulator during SFT data generation.
    \item \textbf{RL rollout patient simulation.} Section~\ref{sec:rl_environment} documents the use of \texttt{Qwen3-30B-A3B} as the patient simulator during DAPO rollouts.
    \item \textbf{Diagnosis judging.} Section~\ref{sec:reward} and Section~\ref{sec:evaluation_pipeline} document the use of \texttt{gpt-4.1-mini} as an LLM judge that extracts condition sets and counts matches.
\end{itemize}

\end{document}